\documentclass[10pt,twocolumn,letterpaper]{article}

\usepackage[pagenumbers]{wacv} % To force page numbers, e.g. for an arXiv version

\usepackage{multicol}
\usepackage{multirow}
\usepackage{booktabs}
\usepackage{amssymb}
\usepackage{tcolorbox}
\usepackage{algorithm}
\usepackage{algpseudocode}
\usepackage{graphicx}
\usepackage{wrapfig}
\usepackage[margin=1in]{geometry}
\tcbuselibrary{skins,breakable}
\usepackage{xcolor}
\usepackage{placeins}
\usepackage{longtable}

\algrenewcommand{\algorithmiccomment}[1]{%
  \hfill{\footnotesize\textcolor{gray!70!black}{$\triangleright$~#1}}}
\algrenewcommand{\alglinenumber}[1]{}

\definecolor{wacvblue}{rgb}{0.21,0.49,0.74}
\usepackage[breaklinks,colorlinks,allcolors=wacvblue]{hyperref}

\def\wacvPaperID{1939}
    \def\confName{WACV}
\def\confYear{2027}

\title{C-GAP: Class‑Aware and Online Prompting Improves Vision–Language Models on Imbalanced Classes}

\author{
    Francis Fernandez \\
    San Diego State University \\
    {\tt\small fafernandez@sdsu.edu}
    \and
    Arash Jahangiri \\
    San Diego State University \\
    {\tt\small ajahangiri@sdsu.edu}
    \and
    Salimeh Sekeh \\
    San Diego State University \\
    {\tt\small ssekeh@sdsu.edu}
}

\begin{document}
\maketitle
\begin{abstract}

Safety-critical perception systems must reliably detect rare object classes within small label spaces, a setting that long-tailed detection methods, designed for hundreds of classes with dense annotation, fundamentally do not address. 
Open-vocabulary detectors offer a promising alternative, as they use natural language queries at inference time, making prompt quality a first-class lever for detection performance. 
We exploit this property to address class imbalance: rather than retraining models or collecting additional annotations, we ask whether iteratively refining the language prompts, fed to frozen detectors, can improve minority-class detection. 
We introduce \textbf{C-GAP} (\underline{C}aption-\underline{G}uided \underline{A}ugmentation and \underline{P}rompting), a detector-agnostic, annotation-free framework that operates in two phases. 
First, we establish a composite caption baseline combining per-image scene descriptions with class-quantity context, which we show outperforms scene-description-only or class-quantity-only prompts across
multiple open-vocabulary architectures and benchmarks. 
%four open-vocabulary architectures (Grounding DINO, OmDet-Turbo, OWLv2, and YOLO-World) and three datasets (MS-COCO, Cityscapes, and a traffic-surveillance benchmark). 
Second, an LLM iteratively refines each image's caption individually, with trials triaged into \textit{accept}, \textit{tentative}, or \textit{regenerate} buckets based on minority-class AP@0.5 against a dynamic threshold derived from the composite baseline. 
Refinement terminates early once sufficient AP@0.5 gain is achieved. 
No detector weights are updated at any stage. Our experiments shows that C-GAP improves minority-class average precision up to 53\% over the baselines. 
On COCO, C-GAP improves minority-class AP@0.5 by ${\sim}81\%$ relative over the composite baseline ($17.69 \rightarrow 32.09$). 
%Results across all four backbones and datasets are reported in the paper. 
Experiments confirm that composite captions provide the critical foundation for effective refinement: using scene-description-only or class-quantity-only prompts as the refinement starting point yields diminishing returns, supporting both stages of C-GAP as necessary contributions.
\end{abstract}    
\begin{figure}[h!]
    \centering
    \includegraphics[width=0.9\linewidth]{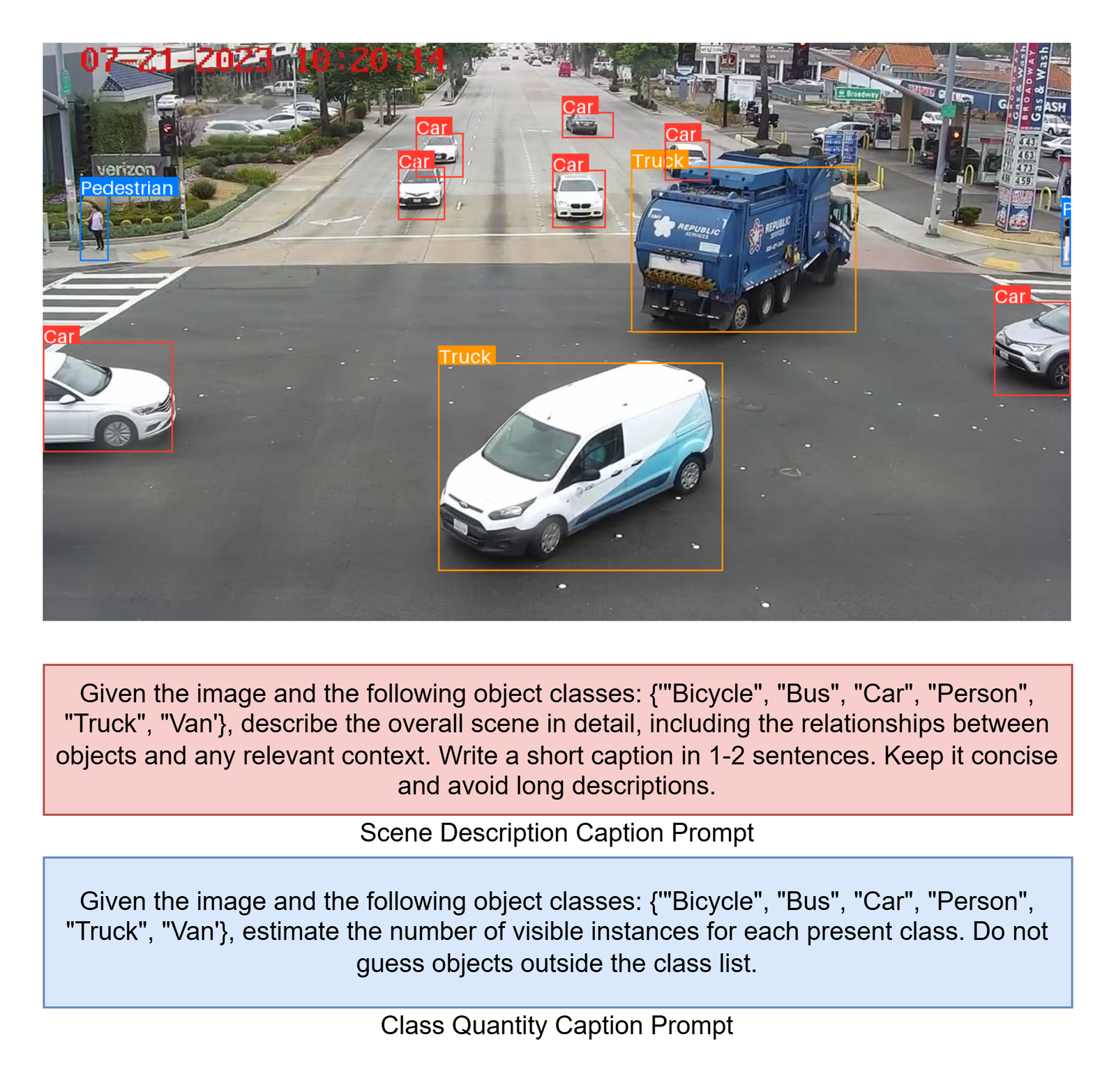}
    \caption{Chula Vista intersection: cars dominate (${\sim}80\%$) while cyclists are severely underrepresented (${\sim}8\%$), motivating minority-class prompt refinement.}
    \label{fig:chula-vista-scene}
\end{figure}

\section{Introduction}
\label{sec:intro}

\begin{figure*}
    \centering
    \includegraphics[width=0.8\linewidth]{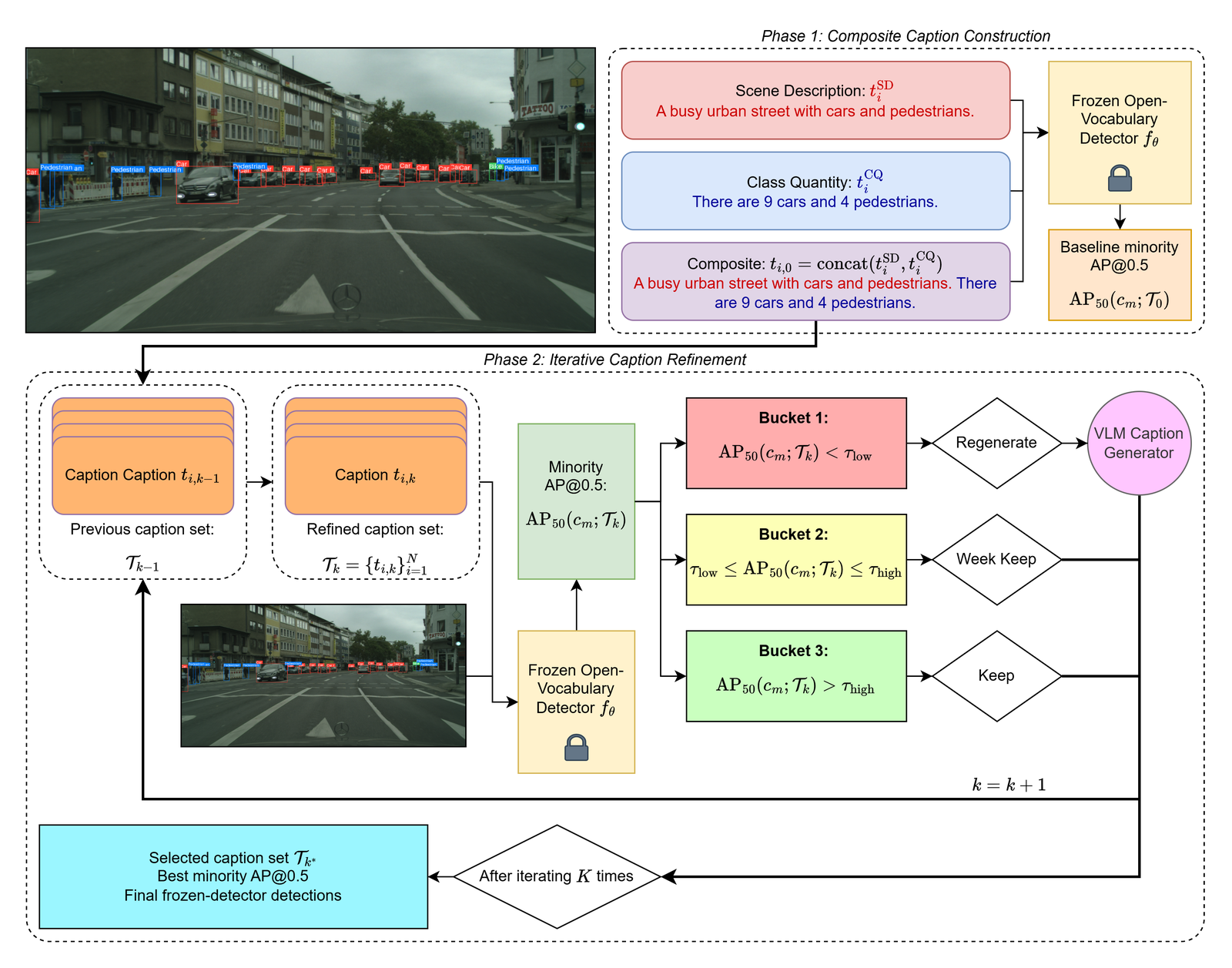}
    \caption{Overview of C-GAP.
    \textbf{Phase~I} (top): per-image Scene Description ($t_i^{\mathrm{SD}}$) and Class Quantity ($t_i^{\mathrm{CQ}}$) captions are generated offline and concatenated into a Composite Caption $t_{i,0}=\operatorname{concat}(t_i^{\mathrm{SD}},t_i^{\mathrm{CQ}})$, forming the initial caption set $\mathcal{T}_0$.
    \textbf{Phase~II} (bottom): a VLM refines $\mathcal{T}_0$ over $K$ trials. Each candidate set $\mathcal{T}_k$ is evaluated by the frozen detector $f_\theta$, yielding minority-class $\mathrm{AP}_{50}(c_m;\mathcal{T}_k)$.
    Trials are triaged into $B_1$ (regenerate, $\mathrm{AP}_{50}<\tau_{\mathrm{low}}$), $B_2$ (tentative, $\tau_{\mathrm{low}}\le\mathrm{AP}_{50}\le\tau_{\mathrm{high}}$), or $B_3$ (keep, $\mathrm{AP}_{50}>\tau_{\mathrm{high}}$).
    The output is $\mathcal{T}_{k^*}$ where $k^*=\arg\max_{0\le k\le K}\mathrm{AP}_{50}(c_m;\mathcal{T}_k)$.
    Detector weights $\theta$ are never updated.}
    \label{fig:cgap_overview}
\end{figure*}

Safety-critical perception systems must reliably detect rare object classes within small label space, a setting where long-tailed detection methods, designed for hundreds of categories with dense annotation, offer limited guidance~\cite{oksuz2020imbalance,gupta2019lvis}.
In urban traffic monitoring, minority classes such as cyclists and pedestrians account for a small fraction of instances yet are the most consequential for collision avoidance and regulatory compliance.
Standard resampling and loss-reweighting remedies require labeled minority instances that may simply not exist in sufficient quantity, and collecting additional annotations is expensive.

Because open-vocabulary detectors (OVDs)~\cite{zareian2021ovrcnn} accept free-form text queries at inference time rather than a fixed classification head, \emph{prompt quality} becomes a direct lever for detection performance, requiring no annotation, no architectural change, and no retraining.
We exploit this property to ask: \emph{can caption design alone improve minority-class detection under low-cardinality class imbalance?}

Captions are typically written as scene descriptions~\cite{chen2015cococaptions}, emphasizing global context and object relationships, which under-emphasizes
minority classes visually dominated by majority ones.
We therefore compare three prompt types, scene description (SD), class quantity (CQ), and their combination (CC), across four open-vocabulary architectures~\cite{liu2023groundingdino,zhao2024omdetturbo,minderer2023owlv2,cheng2024yoloworld} and three datasets~\cite{lin2014coco,cordts2016cityscapes}, finding that composite captions consistently outperform either component alone, with the gap widest for minority classes.
Building on this, we introduce \textbf{C-GAP} (\underline{\textbf{C}}aption-\underline{\textbf{G}}uided \underline{\textbf{A}}ugmentation and \underline{\textbf{P}}rompting): an iterative framework that uses VLM-generated per-image caption refinements, guided by detector-measured minority-class AP@0.5, to push beyond the composite caption baseline without updating any detector weights (Figure~\ref{fig:cgap_overview}).
On COCO (minority class: bus), C-GAP achieves up to ${\sim}81\%$ relative AP@0.5 improvement over the composite baseline across multiple seeds and backbones.

\noindent\textbf{Application: Smart Intersection Monitoring.}
Figure~\ref{fig:chula-vista-scene} shows a fixed-camera intersection where cars dominate (${\sim}80\%$) and cyclists are severely underrepresented (${\sim}8\%$).
Cameras biased toward majority classes risk missing cyclists and pedestrians precisely when they are most vulnerable.
This imbalance compounds as the model continues to encounter car-dominated scenes.
C-GAP addresses this directly: by iteratively refining text prompts using detector AP@0.5 as feedback, it recovers minority-class detections from a frozen detector with no additional labels or retraining.

\noindent\textbf{Contributions:}
\begin{itemize}[leftmargin=*, nosep]
  \item A systematic study of caption-type effects on minority-class detection across four open-vocabulary  backbones and three datasets, comparing CC, SD, or CQ to find  no single static caption type dominates across backbones and datase for underrepresented classes.
  \item The C-GAP framework: an annotation-free, training-free iterative caption refinement pipeline driven by frozen-detector AP@0.5 feedback, with a three-bucket triage mechanism supporting per-image LLM refinement and early termination.
  \item Ablations confirming that composite captions are the necessary foundation for effective refinement, with mAP@0.5 analysis verifying minority-class gains do not degrade majority-class detection.
\end{itemize}

\section{Related Work}
\label{sec:related}

Long-tail detection remedies, such as repeat factor sampling~\cite{gupta2019lvis}, focal loss~\cite{lin2017focal}, class-balanced re-weighting~\cite{cui2019classbalanced}, equalization loss~\cite{tan2020equalization}, and seesaw loss~\cite{wang2021seesaw}, all require sufficient labeled minority instances, a condition that fails in low-cardinality settings~\cite{oksuz2020imbalance}.
Open-vocabulary detection~\cite{zareian2021ovrcnn} breaks this constraint via CLIP distillation~\cite{radford2021clip,gu2021vild}, region-text alignment~\cite{li2022glip,zhong2022regionclip,kamath2021mdetr}, and web-scale self-training~\cite{minderer2023owlv2,bangalath2022bridging,zhou2022detic}, making prompt content a direct performance lever.
Yet prior work on OVDs targets breadth across many categories~\cite{shen2024ape,zang2022ovdetr}, not minority-class precision within a small fixed label space.
Prompt engineering has pursued learned soft embeddings~\cite{zhou2022coop,zhou2022cocoop}, image-conditional prompts~\cite{khattak2023maple}, and detection-specific tuning~\cite{du2022detpro,feng2022promptdet}, all requiring labeled data.
Closest to C-GAP, Cho~\etal~\cite{cho2023pseudocaption} generate offline pseudo-label captions for OVD, while test-time prompt adaptation~\cite{shu2022testtime} refines prompts from per-sample confidence.
C-GAP extends both: rather than one-shot captions or per-sample signals, we iteratively refine per-image prompts using aggregate minority-class AP@0.5 from a frozen detector, with a three-bucket triage requiring no labels, no embedding optimization, and no parameter updates.
\section{Methodology}
\label{sec:method}

\subsection{Problem Formulation}
\label{sec:formulation}

Let $\mathcal{D}=\{(I_i,Y_i)\}_{i=1}^{N}$ denote an object-detection dataset, where $I_i$ is an image and $Y_i=\{(b_{ij},c_{ij})\}_{j=1}^{n_i}$ contains ground-truth boxes and labels from a fixed class set $\mathcal{C}$. We focus on a dataset-specific minority class $c_m\in\mathcal{C}$, whose instances are substantially less frequent than the dominant classes. C-GAP assumes a frozen open-vocabulary detector $f_{\theta}$ with fixed parameters $\theta$. Given an image $I_i$ and a text prompt $t_{i,k}$, the detector outputs predictions $ \hat{Y}_{i,k}=f_{\theta}(I_i,t_{i,k};\mathcal{C})$,
% \begin{equation}
%     \hat{Y}_{i,k}=f_{\theta}(I_i,t_{i,k};\mathcal{C}),
% \end{equation}
where $k$ denotes the caption-refinement trial. Unlike detector training or fine-tuning, C-GAP does not update $\theta$; it only changes the per-image prompts supplied to the detector.

For each trial $k$, C-GAP evaluates a per-image caption set $ \mathcal{T}_k=\{t_{i,k}\}_{i=1}^{N}$.
% \begin{equation}
%     \mathcal{T}_k=\{t_{i,k}\}_{i=1}^{N}.
% \end{equation}

The primary objective is to select the caption set that maximizes minority-class detection:
\begin{equation}
    k^*=
    \arg\max_{0\le k\le K}
    \mathrm{AP}_{50}(c_m;\mathcal{T}_k),
    \label{eq:cgap-objective}
\end{equation}
where $k=0$ denotes the initial caption set and $K$ is the maximum number of refinement trials. We additionally report $\mathrm{mAP}_{50}(\mathcal{C};\mathcal{T}_k)$ as a guardrail to measure whether minority-class gains preserve overall detection quality.

C-GAP consists of two Phases: Phase I constructs a composition of scene description and class-aware captions while Phase II iteratively categorizes captions into buckets and refines the selective caption sets.

\subsection{Phase I: Composite Caption Construction}
\label{sec:phase1}

Given image $I_i$ and class set $C$, we generate two complementary
per-image captions \emph{offline} using an instruction-tuned
vision-language model~$G_\phi$.

\noindent\textbf{Scene Description (SD).} $G_\phi$ describes the
visual scene in natural language grounded to the target classes,
producing caption $t^{\mathrm{SD}}_i$ containing contextual and
relational cues.

\noindent\textbf{Class Quantity (CQ).} $G_\phi$ estimates the
visible count of each class in $C$, producing caption
$t^{\mathrm{CQ}}_i$ that explicitly signals class presence and
approximate frequency.

\noindent\textbf{Composite Caption (CC).} The two captions are
concatenated to form the initial per-image prompt:
\begin{equation}
  t_{i,0} = \mathrm{concat}\!\left(t^{\mathrm{SD}}_i,\,t^{\mathrm{CQ}}_i\right).
  \label{eq:composite}
\end{equation}
The initial caption set is $\mathcal{T}_0 = \{t_{i,0}\}_{i=1}^N$.
Because SD and CQ encode complementary signals, scene context
and class salience respectively, their combination provides a
richer per-image initialization than either alone, as confirmed
empirically in Section~\ref{sec:experiments}.

% \subsection{Phase I: Composite Caption Construction}
% \label{sec:phase1}

% Given image $I_i$ and class set $\mathcal{C}$, we generate two complementary per-image captions offline using an instruction-tuned vision-language model (VLM).

% \noindent\textbf{Scene Description (SD).}
% The model describes the visual scene in natural language, grounded to the target classes. This produces a scene-description caption $t_i^{\mathrm{SD}}$ containing contextual and relational cues.

% \noindent\textbf{Class Quantity (CQ).}
% The model estimates the visible count of each target class in $\mathcal{C}$. This produces a class-quantity caption $t_i^{\mathrm{CQ}}$ that explicitly signals class presence and approximate frequency.

% \noindent\textbf{Composite Caption (CC).}
% We concatenate the two captions to form the initial per-image prompt
% \begin{equation}
%     t_{i,0}=
%     \operatorname{concat}\left(t_i^{\mathrm{SD}},t_i^{\mathrm{CQ}}\right).
%     \label{eq:composite}
% \end{equation}
% The initial caption set is $\mathcal{T}_0=\{t_{i,0}\}_{i=1}^{N}$.

\subsection{Phase II: Iterative Caption Refinement}
\label{sec:phase2}

Starting from $\mathcal{T}_0$, C-GAP refines per-image captions iteratively using minority-class AP@0.5 as the feedback signal.
At each trial $k$, the caption generator $G_\phi$ produces a new caption set $\mathcal{T}_k = \{t_{i,k}\}_{i=1}^N$ for every image independently, conditioned on that image's own previous caption and the aggregate AP@0.5 feedback from trial $k{-}1$.
The frozen detector then evaluates $\mathcal{T}_k$.

\noindent\textbf{Bucket Triage.}
Each candidate caption set is assigned to one of three buckets using dynamic thresholds around the composite-caption baseline:
\begin{equation}
\mathrm{bucket} =
\begin{cases}
  \text{regenerate} & \text{if } \mathrm{AP}_{50}(c_m;\mathcal{T}_k) < \tau_{\mathrm{low}}, \\
  \text{tentative}  & \text{if } \tau_{\mathrm{low}} \le \mathrm{AP}_{50}(c_m;\mathcal{T}_k) \le \tau_{\mathrm{high}}, \\
  \text{keep}       & \text{if } \mathrm{AP}_{50}(c_m;\mathcal{T}_k) > \tau_{\mathrm{high}},
\end{cases}
\label{eq:bucket}
\end{equation}
where $\tau_{\mathrm{high}} = \mathrm{AP}_{\mathrm{CC}} + \delta$
and $\tau_{\mathrm{low}} = \mathrm{AP}_{\mathrm{CC}} - \delta$
for margin $\delta$.
B1 (regenerate) strengthens corrective feedback for the next trial.
B2 (tentative) weakly retains the candidate and continues.
B3 (keep) may terminate refinement early.
The full procedure is given in Algorithm~\ref{alg:cgap}.

% \subsection{Phase II: Iterative Caption Refinement}
% \label{sec:phase2}

% Starting from $\mathcal{T}_0$, C-GAP refines the per-image caption iteratively using detector-measured AP@0.5 on the minority class as a feedback signal.
% At each trial $k$, a caption generator produces a new caption set $\mathcal{T}_k=\{t_{i,k}\}_{i=1}^{N}$ for every image independently, conditioned on that image's own previous caption and aggregate ap@50 feedback from trial $k-1$.
% The frozen detector is then evaluated using $\mathcal{T}_k$.

% \noindent\textbf{Bucket Triage.}
% Each candidate caption set is assigned to a bucket using dynamic thresholds around the initial composite-caption baseline:
% \begin{equation}
%     \text{bucket} =
%     \begin{cases}
%         \textit{regenerate} & \text{if } \mathrm{AP}_{50}(c_m;\mathcal{T}_k) < \tau_{\mathrm{low}}, \\
%         \textit{tentative}  & \text{if } \tau_{\mathrm{low}} \le \mathrm{AP}_{50}(c_m;\mathcal{T}_k) \le \tau_{\mathrm{high}}, \\
%         \textit{keep}       & \text{if } \mathrm{AP}_{50}(c_m;\mathcal{T}_k) > \tau_{\mathrm{high}}, 
%     \end{cases}
%     \label{eq:buckets}
% \end{equation}
% where $\tau_{\mathrm{high}} = \mathrm{AP}_{\mathrm{CC}} + \delta$ and $\tau_{\mathrm{low}} = \mathrm{AP}_{\mathrm{CC}} - \delta$ for margin $\delta = 0.05$.
% The bucket decision controls the feedback supplied to the next refinement trial: $B_1$ strengthens corrective regeneration, $B_2$ weakly retains the candidate, and $B_3$ may terminate refinement early. The full procedure is given in Algorithm~\ref{alg:cgap}.

\begin{algorithm}[t]
    \caption{C-GAP: Caption-Guided Augmentation and Prompting}
    \label{alg:cgap}
    \footnotesize
    \begin{algorithmic}
        \Statex \textbf{Input:}\enspace
        image set $\{I_i\}_{i=1}^{N}$;\enspace
        class set $\mathcal{C}$;\enspace
        minority class $c_m$;\enspace
        frozen detector $f_{\theta}$;\enspace
        caption generator $G_{\phi}$;\enspace
        margin $\delta$;\enspace
        max trials $K$
        \Statex \textbf{Output:}\enspace
        best caption set $\mathcal{T}_{k^*}$ and detections $\hat{\mathcal{Y}}_{k^*}$
    \end{algorithmic}

    \vspace{4pt}
    \begin{tcolorbox}[
        enhanced,
        colback=cyan!8!white,
        colframe=cyan!50!blue,
        colbacktitle=cyan!50!blue,
        coltitle=white,
        boxrule=0.8pt, arc=3pt, boxsep=0pt,
        left=4pt, right=4pt, top=3pt, bottom=3pt,
        toptitle=2pt, bottomtitle=2pt,
        title={\footnotesize\bfseries Phase I \quad Composite Caption Construction},
        fonttitle=\bfseries]
        \begin{algorithmic}
            \For{each image $I_i$}
            \State $t_i^{\mathrm{SD}} \leftarrow
            G_{\phi}(I_i,\mathcal{C},\texttt{SD-prompt})$
            \State $t_i^{\mathrm{CQ}} \leftarrow
            G_{\phi}(I_i,\mathcal{C},\texttt{CQ-prompt})$
            \State $t_{i,0} \leftarrow
            \operatorname{concat}(t_i^{\mathrm{SD}},t_i^{\mathrm{CQ}})$
            \Comment{composite caption}
            \EndFor
            \State $\mathcal{T}_0 \leftarrow \{t_{i,0}\}_{i=1}^{N}$
            \State $\hat{\mathcal{Y}}_0 \leftarrow
            f_{\theta}(\{I_i\}_{i=1}^{N},\mathcal{T}_0;\mathcal{C})$
            \Comment{frozen detector}
            \State $a_0 \leftarrow \mathrm{AP}_{50}(c_m;\mathcal{T}_0)$
            \State $\tau_{\mathrm{low}} \leftarrow a_0-\delta$;\quad
            $\tau_{\mathrm{high}} \leftarrow a_0+\delta$
            \State $k^* \leftarrow 0$;\quad
            $a^* \leftarrow a_0$;\quad
            $\beta_0 \leftarrow \textit{regenerate}$
        \end{algorithmic}
    \end{tcolorbox}

    \vspace{3pt}
    \begin{tcolorbox}[
        enhanced,
        colback=orange!6!white,
        colframe=orange!70!red!80,
        colbacktitle=orange!70!red!80,
        coltitle=white,
        boxrule=0.8pt, arc=3pt, boxsep=0pt,
        left=4pt, right=4pt, top=3pt, bottom=3pt,
        toptitle=2pt, bottomtitle=2pt,
        title={\footnotesize\bfseries Phase II \quad Iterative Per-Image Refinement},
        fonttitle=\bfseries]
        \begin{algorithmic}
            \For{$k=1,\ldots,K$}
            \For{each image $I_i$}
            \State $t_{i,k} \leftarrow
            G_{\phi}(t_{i,0},t_{i,k-1},c_m,\mathcal{C},a_{k-1},\beta_{k-1})$
            \Comment{per-image refinement}
            \EndFor
            \State $\mathcal{T}_k \leftarrow \{t_{i,k}\}_{i=1}^{N}$
            \State $\hat{\mathcal{Y}}_k \leftarrow
            f_{\theta}(\{I_i\}_{i=1}^{N},\mathcal{T}_k;\mathcal{C})$
            \Comment{no weight update}
            \State $a_k \leftarrow \mathrm{AP}_{50}(c_m;\mathcal{T}_k)$
            \If{$a_k>a^*$}
            \State $k^* \leftarrow k$;\quad $a^* \leftarrow a_k$
            \EndIf

            \vspace{3pt}
            \begin{tcolorbox}[
                enhanced,
                colback=green!6!white,
                colframe=green!45!black,
                colbacktitle=green!45!black,
                coltitle=white,
                boxrule=0.7pt, arc=3pt, boxsep=0pt,
                left=4pt, right=4pt, top=3pt, bottom=3pt,
                toptitle=2pt, bottomtitle=2pt,
                title={\footnotesize\bfseries Bucket Triage},
                fonttitle=\bfseries]
                \If{$a_k>\tau_{\mathrm{high}}$}
                    \State $\beta_k \leftarrow B_3;(\textit{keep})$
                    \State \textbf{break} \Comment{early stop}
                \ElsIf{$\tau_{\mathrm{low}}\le a_k\le\tau_{\mathrm{high}}$}
                    \State $\beta_k \leftarrow B_2;(\textit{weak keep})$
                    \Comment{retain candidate; continue}
                \Else
                    \State $\beta_k \leftarrow B_1;(\textit{regenerate})$
                    \Comment{discard candidate; strengthen feedback}
                \EndIf
            \end{tcolorbox}
            \EndFor
        \end{algorithmic}
    \end{tcolorbox}

    \vspace{3pt}
    \begin{algorithmic}
    \State \Return $\mathcal{T}_{k^*},\hat{\mathcal{Y}}_{k^*}$
    \end{algorithmic}
\end{algorithm}
\section{Experiments}
\label{sec:experiments}

%\subsection{Datasets}

{\bf Datasets:} We evaluate on three datasets covering standard benchmarks and real-world deployment, each with a small fixed label space and a distinct minority class. Because we are not training our models, we are only using the validation set for each dataset.

\noindent\textbf{MS-COCO}~\cite{lin2014coco}.
2017 validation split filtered to \{\textit{person, bicycle, car, bus, truck}\}, yielding 2{,}948 images, and bus is the minority class.

\noindent\textbf{Cityscapes}~\cite{cordts2016cityscapes}.
500-image validation split, and polygon annotations converted to
axis-aligned bounding boxes.
\textit{rider} is merged into \textit{person}, and truck is the minority class.

\noindent\textbf{Chula Vista.}
288 images from fixed-position smart-intersection cameras in Chula Vista, California. 
It includes six classes \{\textit{Bike, Bus, Car, Pedestrian, Truck, Van}\}.
Bike is the minority class (${\sim}8\%$ of instances vs.\ ${\sim}80\%$ for Car), the most severe imbalance of the three
datasets.
Per-class instance counts are in the supplementary material.

%\subsection{Detector Backbones}

\noindent{\bf Detector Backbones:} 
We evaluate four open-vocabulary architectures spanning distinct design families, all used \textbf{zero-shot with frozen weights}, and no fine-tuning is performed at any stage.
\textbf{Grounding DINO}~\cite{liu2023groundingdino} (BERT-based deep cross-modal fusion),
\textbf{OmDet-Turbo}~\cite{zhao2024omdetturbo} (RT-DETR with efficient fusion head),
\textbf{OWLv2}~\cite{minderer2023owlv2} (ViT with late-fusion text conditioning), and
\textbf{YOLO-World}~\cite{cheng2024yoloworld} (single-stage CLIP vocabulary re-parameterization).
The implementation details are provided in the supplementary materials. 

%---------------Salimeh: moved this section to the SM--------
% \subsection{Implementation Details}

\noindent {\bf Evaluation Metrics: }
Minority-class AP@0.5 (bus/COCO, truck/Cityscapes, Bike/Chula Vista) is the primary metric, following Eq.~\ref{eq:cgap-objective}.
All-class mAP@0.5 is reported as a guardrail against majority-class degradation. 
All metrics use the standard COCO evaluation protocol (\texttt{pycocotools}). 
For C-GAP, each seed first evaluates the initial CC caption set and then evaluates each refined caption set up to the trial budget $K$. 
The reported score is the best found across all trials, including the initial CC baseline as trial $k=0$.
So C-GAP results represent best-over-budget prompt selection averaged over three seeds.

\section{Results and Analysis}
\label{sec:results}

\subsection{Caption Type Study}

Table~\ref{tab:caption-type} reports minority-class AP@0.5 for SD, CQ,
and CC as static detector prompts ($4{\times}3{\times}3{=}36$
configurations, llava-phi3, no refinement).
No single static type dominates: the best prompt varies by backbone,
dataset, and class, with gaps as large as 11\,pp between caption types
for the same configuration (YOLO-World/COCO: CQ\,24.47 vs.\ CC\,13.43).
Grounding DINO and OmDet-Turbo score 0.00 minority AP@0.5 across all
caption types on Chula Vista, reflecting the extreme scarcity of Bike
instances and demonstrating exactly the failure mode C-GAP is designed
to address.
This backbone- and dataset-specific sensitivity confirms that static
prompt selection is brittle: the best caption type must be discovered
per backbone, which in practice requires labeled validation data.
C-GAP removes that requirement by starting from CC as a backbone-agnostic
initialization and using detector-measured minority AP@0.5 to iteratively
drive per-image refinement beyond any fixed static baseline.

\begin{table}[t]
\centering
\caption{Minority-class AP@0.5 (\%) per static caption type (llava-phi3,
  no refinement). Best per backbone--dataset in \textbf{bold}.
  Minority: bus/COCO, truck/Cityscapes, Bike/Chula Vista.}
\label{tab:caption-type}
\resizebox{\columnwidth}{!}{%
\begin{tabular}{llccc}
\toprule
Model & Type & COCO & Cityscapes & Chula Vista \\
\midrule
\multirow{3}{*}{Grounding DINO}
  & SD & \textbf{33.18} & 2.55 & 0.00 \\
  & CQ & 26.50 & \textbf{5.85} & 0.00 \\
  & CC & 26.61 & 4.92 & 0.00 \\
\midrule
\multirow{3}{*}{OmDet-Turbo}
  & SD & \textbf{64.68} & 6.82 & 0.00 \\
  & CQ & 60.34 & 5.39 & \textbf{5.61} \\
  & CC & 62.19 & \textbf{7.90} & 0.00 \\
\midrule
\multirow{3}{*}{OWLv2}
  & SD & 84.48 & 19.13 & 0.40 \\
  & CQ & \textbf{85.96} & \textbf{19.14} & \textbf{0.46} \\
  & CC & 83.97 & 19.05 & 0.44 \\
\midrule
\multirow{3}{*}{YOLO-World}
  & SD & 18.87 & \textbf{10.95} & \textbf{11.65} \\
  & CQ & \textbf{24.47} & 4.94 & 5.61 \\
  & CC & 13.43 & 7.60 & 6.12 \\
\bottomrule
\end{tabular}}
\end{table}

\subsection{C-GAP Refinement Results}

Table~\ref{tab:cgap-results} compares the CC baseline against C-GAP
(llava-phi3, $K{=}15$, mean/3 seeds) across all four backbones and three
datasets.
C-GAP selects the best minority AP@0.5 across all $K$ trials including
$k{=}0$ (the CC baseline), so it cannot score below CC by construction.

\begin{table}[t]
\centering
\caption{Minority AP@0.5 (Min, \%) and overall mAP@0.5 (\%),
  CC vs.\ C-GAP (llava-phi3, $K{=}15$, mean/3 seeds).
  Minority: bus/COCO, truck/Cityscapes, Bike/Chula Vista.}
\label{tab:cgap-results}
\resizebox{\columnwidth}{!}{%
\begin{tabular}{ll|cc|cc|cc}
\toprule
& & \multicolumn{2}{c|}{COCO}
  & \multicolumn{2}{c|}{Cityscapes}
  & \multicolumn{2}{c}{Chula Vista} \\
Model & Method & mAP & Min & mAP & Min & mAP & Min \\
\midrule
\multirow{2}{*}{\shortstack[l]{Grounding\\DINO}}
  & CC    & 17.25 & 15.30 & 10.97 &  8.26 & 10.68 &  0.00 \\
  & C-GAP & \textbf{18.06} & \textbf{23.58}
           &  9.80 & \textbf{12.69}
           & \textbf{15.18} & \textbf{10.29} \\
\midrule
\multirow{2}{*}{\shortstack[l]{OmDet-\\Turbo}}
  & CC    & 40.66 & 53.56 & 23.61 & 21.15 & 25.93 &  0.00 \\
  & C-GAP & 38.54 & \textbf{63.02}
           & 21.16 & \textbf{22.10}
           & 23.06 & \textbf{3.27} \\
\midrule
\multirow{2}{*}{OWLv2}
  & CC    & 56.15 & 75.83 & 52.66 & 44.63 & 33.16 &  0.44 \\
  & C-GAP & 56.13 & \textbf{77.13}
           & \textbf{53.07} & \textbf{44.78}
           & 33.16 &  0.44 \\
\midrule
\multirow{2}{*}{\shortstack[l]{YOLO-\\World}}
  & CC    & 14.22 & 17.69 & 10.07 & 17.27 &  9.72 &  6.12 \\
  & C-GAP & \textbf{24.10} & \textbf{32.09}
           & \textbf{21.28} & \textbf{25.15}
           &  9.72 &  6.12 \\
\bottomrule
\end{tabular}}
\end{table}

\begin{figure*}[t]
\centering
\includegraphics[width=\linewidth]{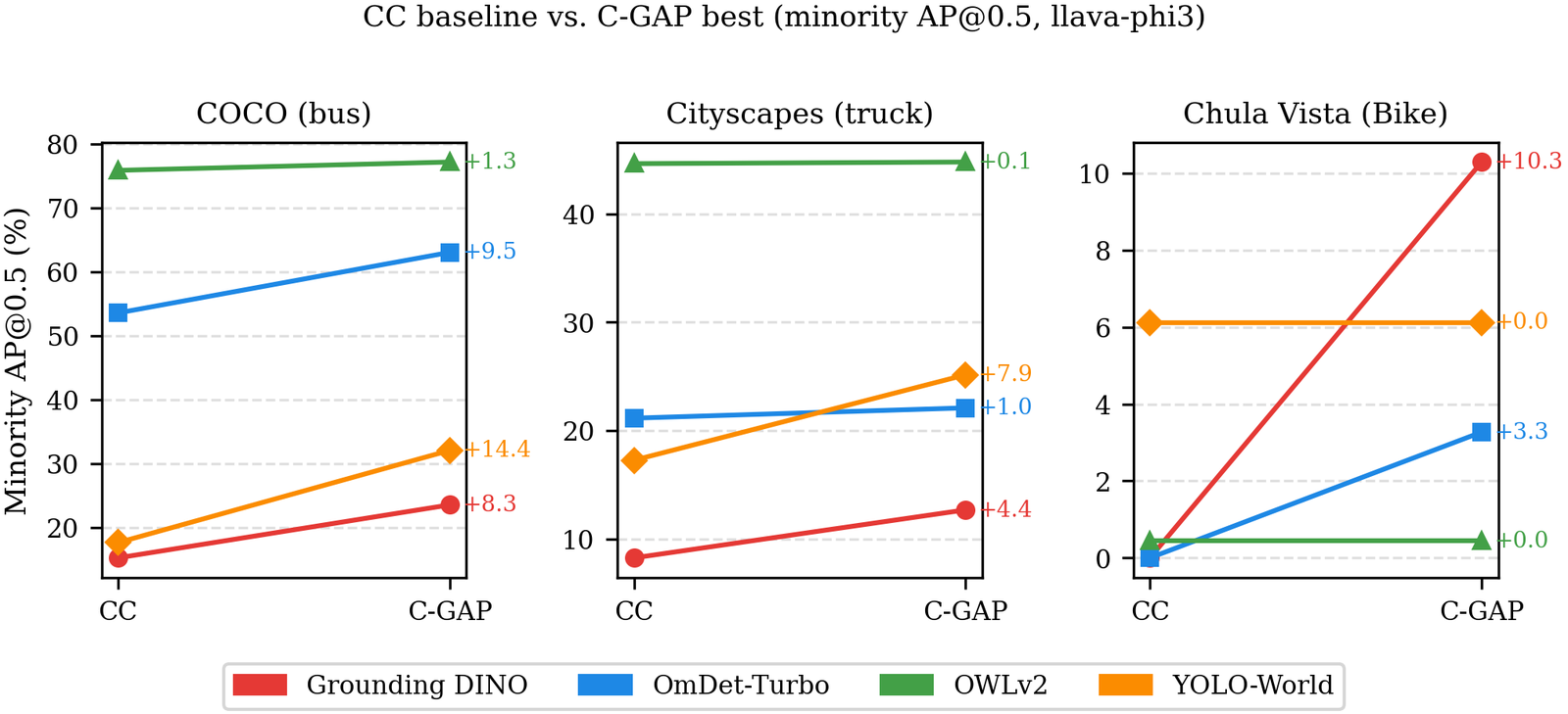}
\caption{Paired CC$\to$C-GAP minority AP@0.5 per backbone and dataset.
  Each line is one frozen detector; upward slopes indicate minority-class
  gains from caption refinement alone.}
\label{fig:paired_slope_cc_vs_cgap}
\end{figure*}

\begin{figure}[t]
\centering
\includegraphics[width=\linewidth]{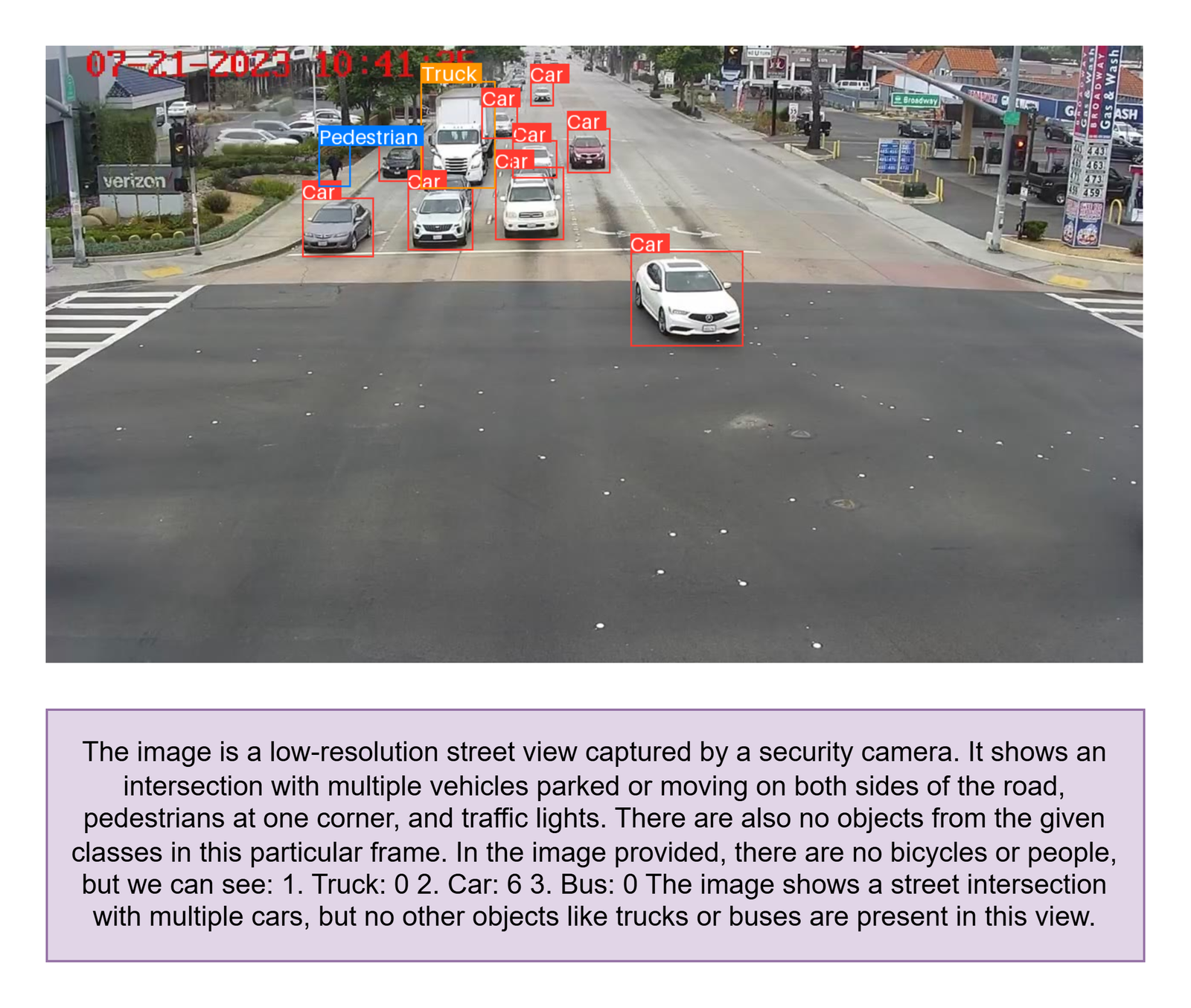}
\caption{An example C-GAP refined caption on OWLv2 with llava-phi3,
  $K=10$. The refined prompt explicitly surfaces the minority class,
  driving the detector toward improved Bike detections.}
\label{fig:chula-vista-caption-ex}
\end{figure}

\begin{figure*}[t]
\centering
\includegraphics[width=\linewidth]{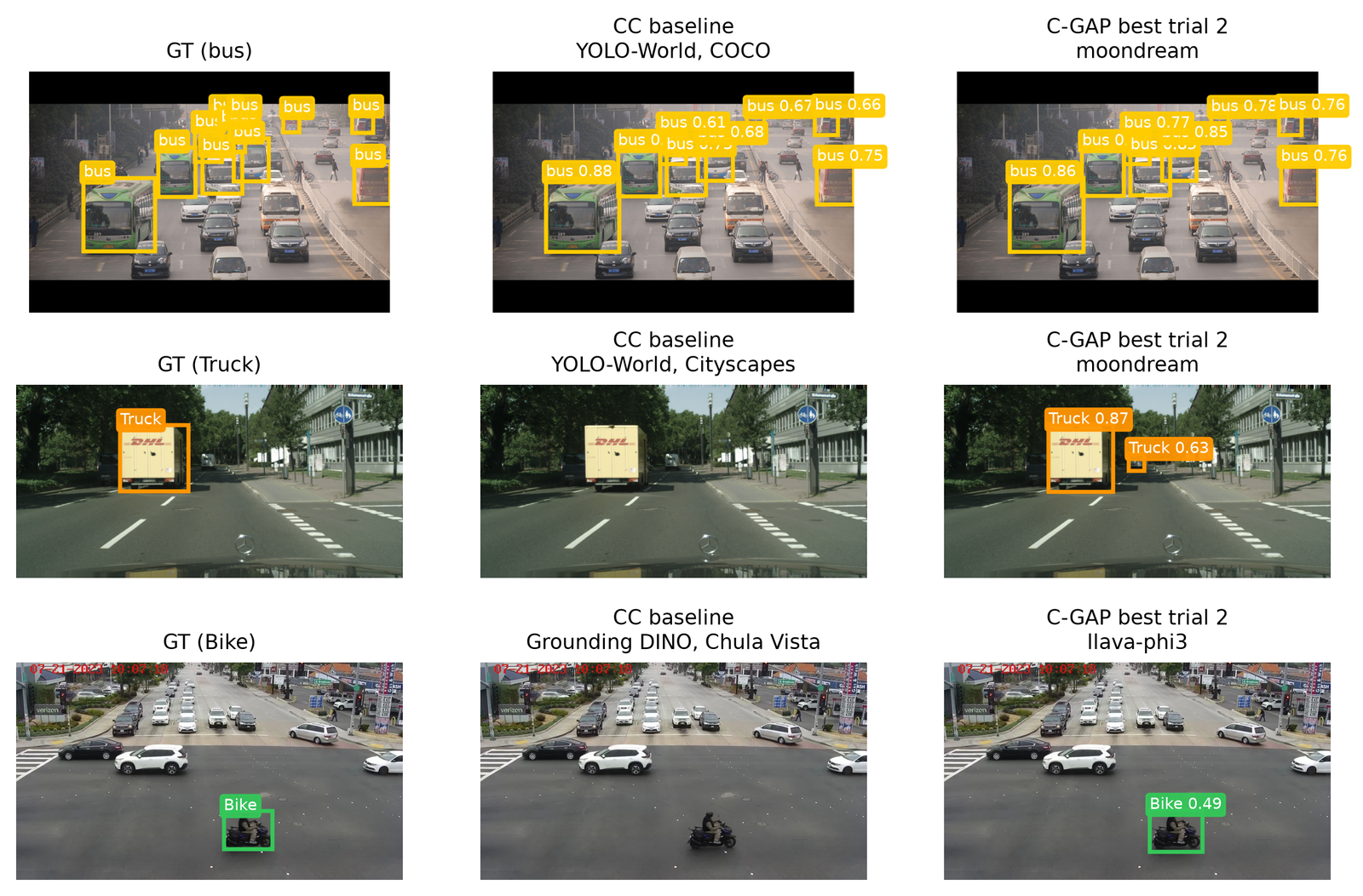}
\caption{Qualitative comparison of CC baseline (top) and C-GAP (bottom).
  C-GAP recovers minority-class detections missed by the baseline using
  the same frozen detector weights.}
\label{fig:qualitative-ex}
\end{figure*}

C-GAP improves minority AP@0.5 in 10 of 12 backbone--dataset pairs.
Grounding DINO improves on all three datasets: COCO bus
$15.30{\to}23.58$ ($+8.28$\,pp), Cityscapes truck $8.26{\to}12.69$
($+4.43$\,pp, ${\sim}54\%$ relative), and Chula Vista Bike
$0.00{\to}10.29$ ($+10.29$\,pp), recovering nonzero minority detections
from a zero-recall CC baseline entirely through caption refinement.
OmDet-Turbo gains $+9.46$\,pp on COCO bus, $+0.95$\,pp on Cityscapes
truck, and $+3.27$\,pp on Chula Vista Bike.
YOLO-World achieves the largest absolute gains: $+14.40$\,pp on COCO and
$+7.88$\,pp on Cityscapes.
OWLv2 on Chula Vista is the only configuration where C-GAP equals CC:
all trials land in B2 and the keep threshold is never reached.
Table~\ref{tab:cgap-minority-gain} summarizes the per-backbone gains.
Figure~\ref{fig:paired_slope_cc_vs_cgap} visualizes the CC-to-C-GAP
trajectory per backbone, Figure~\ref{fig:chula-vista-caption-ex} shows
an example refined caption, and Figure~\ref{fig:qualitative-ex} shows
representative detection outputs.

\begin{table}[t]
\centering
\caption{Minority AP@0.5 gain ($\Delta$\,pp) of C-GAP over CC
  (llava-phi3, $K{=}15$). $0.00$ = C-GAP equals CC by construction.}
\label{tab:cgap-minority-gain}
\resizebox{\columnwidth}{!}{%
\begin{tabular}{lccc}
\toprule
Backbone & COCO & Cityscapes & Chula Vista \\
\midrule
Grounding DINO & $+7.86$  & $+2.21$ & $+5.14$ \\
OmDet-Turbo    & $+6.32$  & $+0.47$ & $+3.04$ \\
OWLv2          & $+1.30$  & $+0.14$ & $0.00$  \\
YOLO-World     & $+21.67$ & $+6.16$ & $0.00$  \\
\bottomrule
\end{tabular}}
\end{table}

\begin{figure}[t]
\centering
\includegraphics[width=\linewidth]{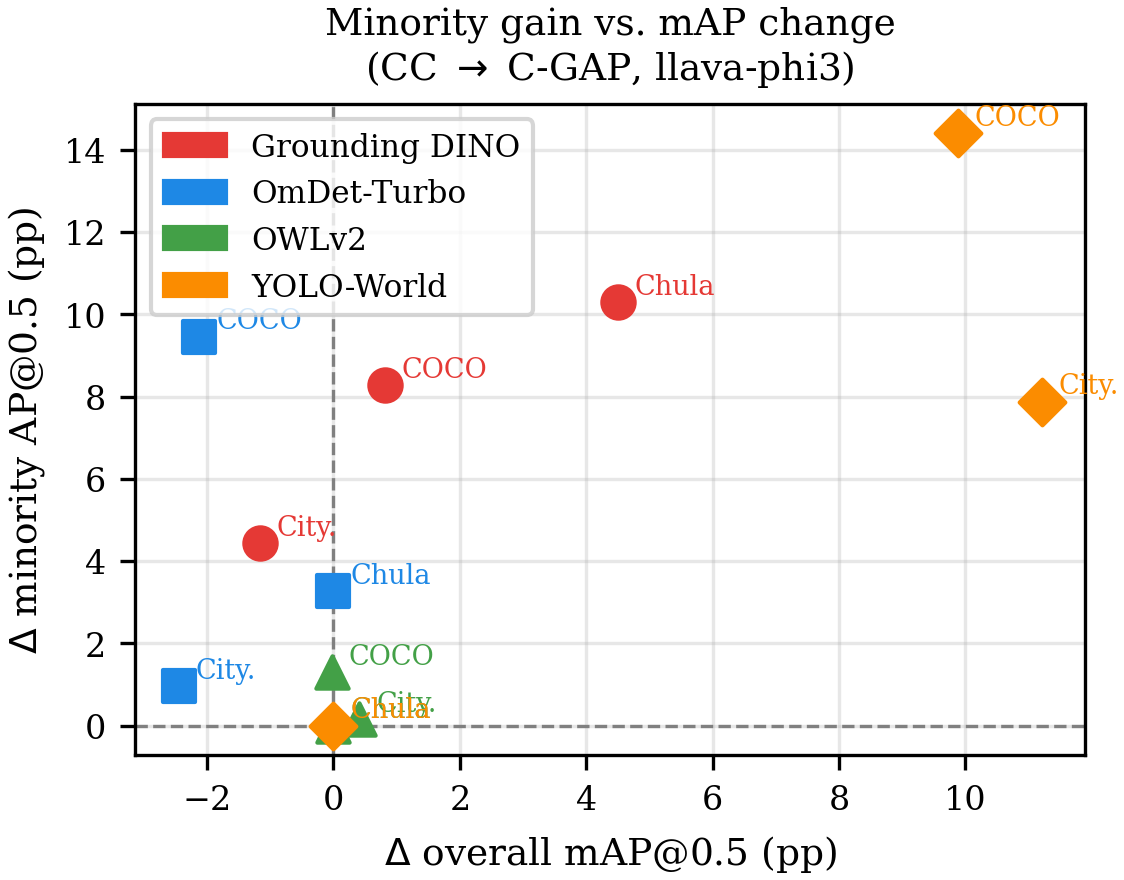}
\caption{Minority AP@0.5 gain vs.\ overall mAP@0.5 change
  (CC$\to$C-GAP). Above $x$-axis: improved minority detection.
  Right of $y$-axis: improved overall detection.}
\label{fig:minority_gain_vs_map_change}
\end{figure}

\noindent\textbf{mAP@0.5 guardrail.}
Overall mAP@0.5 decreases for Grounding DINO on Cityscapes
($10.97{\to}9.80$, $-1.17$\,pp) and OmDet-Turbo on COCO
($40.66{\to}38.54$, $-2.12$\,pp) and Cityscapes ($23.61{\to}21.16$,
$-2.45$\,pp), reflecting a shift in detector attention toward the minority
class at some cost to majority classes.
OWLv2 on COCO demonstrates the ideal outcome: negligible overall mAP@0.5
change ($-0.02$\,pp) alongside a $+1.30$\,pp minority gain.
Figure~\ref{fig:minority_gain_vs_map_change} plots the full tradeoff,
with most pairs above the $x$-axis confirming that minority-class
improvement is the dominant effect.

% \subsection{Ablation Studies}
% \label{sec:ablations}

\subsection{Ablation}
\noindent\textbf{$\mathcal{T}_0$ Initialization}
% \newline
Table~\ref{tab:t0-ablation} compares C-GAP from SD, CQ, and CC starting
points on COCO/bus (Grounding DINO, llava-phi3, $K{=}15$, stratified
working set, mean/3 seeds).
$\sigma$ is the standard deviation of per-seed best-trial indices.
All three initializations benefit from AP-guided refinement, confirming
C-GAP is not restricted to a single starting point.
SD yields the strongest absolute result ($21.53{\to}35.69$, $+14.16$\,pp,
B3\,=\,30\%, $\sigma{=}1.5$), consistent with scene descriptions providing
richer localization context for Grounding DINO's BERT encoder.
CQ improves from $8.09{\to}16.56$ ($+8.47$\,pp, B3\,=\,3\%,
$\sigma{=}1.7$) and CC from $15.30{\to}23.58$ ($+8.28$\,pp, B3\,=\,11\%,
$\sigma{=}6.1$).
CC is used as the primary initialization because it provides
backbone-agnostic, semantically complete starting prompts without
requiring prior knowledge of which component dominates per backbone.

\begin{figure}[t]
\centering
\includegraphics[width=\linewidth]{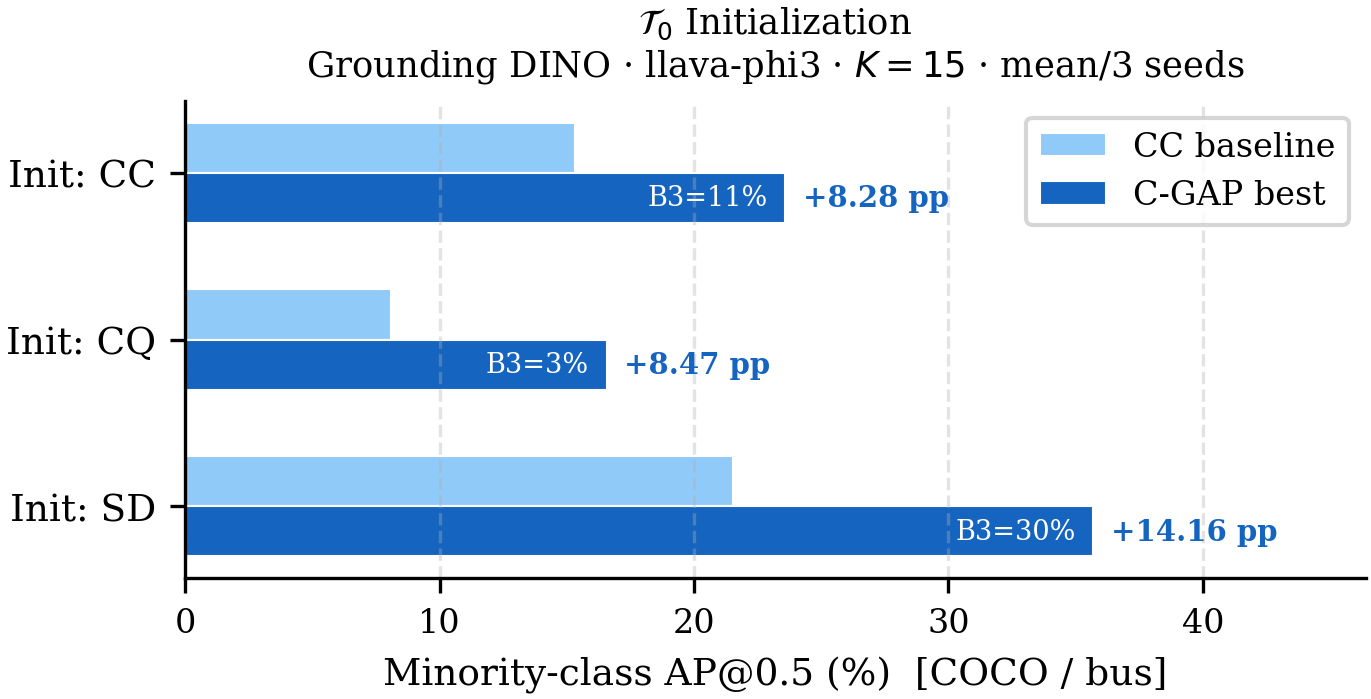}
\caption{$\mathcal{T}_0$ initialization: CC baseline and C-GAP best
  minority AP@0.5 for SD, CQ, and CC starting captions (Grounding DINO,
  COCO/bus, llava-phi3, $K=15$, mean/3 seeds).}
\label{fig:t0_ablation}
\end{figure}

\begin{table}[t]
\centering
\caption{$\mathcal{T}_0$ initialization: COCO/bus, Grounding DINO,
  llava-phi3, $K{=}15$, mean/3 seeds.
  $\sigma$~=~std dev of per-seed best-trial indices.}
\label{tab:t0-ablation}
\begin{tabular}{lrrrrc}
\toprule
$\mathcal{T}_0$ & CC base. & C-GAP & $\Delta$\,pp & B3\% & $\sigma$ \\
\midrule
SD & 21.53 & 35.69 & $+14.16$ & 30 & 1.5 \\
CQ &  8.09 & 16.56 & $+8.47$  &  3 & 1.7 \\
CC & 15.30 & 23.58 & $+8.28$  & 11 & 6.1 \\
\bottomrule
\end{tabular}
\end{table}

\noindent{\bf Phase~II LLM}
Table~\ref{tab:llm-comparison} compares llava-phi3 and moondream as
Phase~II generators (Phase~I captions fixed at llava-phi3, $K{=}15$,
mean over 4 backbones $\times$ 3 seeds).
llava-phi3 outperforms moondream on all three datasets, with the largest
gaps on Cityscapes ($+7.51$\,pp) and Chula Vista ($+3.71$\,pp).
The generator difference, however, is not uniform across backbones.
Table~\ref{tab:generator-buckets} reveals the mechanism: moondream
triggers B1 (regenerate) in 73--96\% of trials for Grounding DINO and
OmDet-Turbo on COCO and Cityscapes, indicating its shorter outputs
consistently fall below $\tau_{\mathrm{low}}$.
Yet on YOLO-World, the pattern inverts: moondream achieves B3\,=\,43\%
on both COCO and Cityscapes vs.\ phi3's 8\% and 0\%, because
moondream's concise outputs avoid the composite-prompt-length penalty
that suppresses phi3 in YOLO-World's vocabulary re-parameterization.
This backbone--generator interaction confirms that generator choice is a
meaningful design variable, not a free parameter.
Full bucket statistics are in the supplementary materials.

\begin{table}[t]
\centering
\caption{Phase~II generator: mean minority AP@0.5 (\%),
  $K{=}15$, mean over 4 backbones $\times$ 3 seeds per dataset.}
\label{tab:llm-comparison}
\resizebox{\columnwidth}{!}{%
\begin{tabular}{lccc}
\toprule
Generator & COCO & Cityscapes & Chula Vista \\
\midrule
llava-phi3 & \textbf{47.39} & \textbf{25.70} & \textbf{6.69} \\
moondream  & 45.02 & 18.19 & 2.98 \\
\bottomrule
\end{tabular}}
\end{table}

\noindent{\bf Trial Budget $K$:}
Table~\ref{tab:k-ablation} reports bucket rates and median best trial for
all backbone--$K$ combinations run on Chula Vista (llava-phi3, 3 seeds).
Three qualitatively distinct triage regimes emerge.
\textbf{OmDet-Turbo} at $K{=}10$ is the most efficient: B3\,=\,43\%,
median best trial~2, $\sigma{=}1.5$.
At $K{=}15$ the B3 rate collapses to 2\% (median~11, $\sigma{=}7.8$),
reflecting over-searching in B2, at $K{=}20$ B3 partially recovers to
13\% (median~3, $\sigma{=}9.5$).
\textbf{Grounding DINO} at $K{=}15$: B3\,=\,15\%, median~5,
$\sigma{=}5.7$.
\textbf{OWLv2} at $K{=}10$: B2\,=\,100\%, so all trials fall between
thresholds without triggering a keep, and C-GAP equals CC.
\textbf{YOLO-World} at $K{=}15,20$: B1\,=\,100\%, all captions
discarded, and C-GAP again equals CC.
$K{=}15$ is the recommended default; $K{=}10$ is more efficient for
OmDet-Turbo on Chula Vista.
Full convergence statistics are in the supplementary materials.

\begin{table}[t]
\centering
\caption{Trial budget $K$ ablation: bucket distribution and convergence
  on Chula Vista (llava-phi3, 3 seeds).
  $\sigma$~=~std dev of per-seed best-trial indices.}
\label{tab:k-ablation}
\resizebox{\columnwidth}{!}{%
\begin{tabular}{llrrrrc}
\toprule
Backbone & $K$ & B1\% & B2\% & B3\% & Med. & $\sigma$ \\
\midrule
G-DINO  & 15 &   0 &  85 &  15 &  5 & 5.7 \\
\midrule
\multirow{3}{*}{OmDet}
        & 10 &   0 &  57 &  43 &  2 & 1.5 \\
        & 15 &   0 &  98 &   2 & 11 & 7.8 \\
        & 20 &   0 &  88 &  13 &  3 & 9.5 \\
\midrule
OWLv2   & 10 &   0 & 100 &   0 &  0 & 0.0 \\
\midrule
\multirow{2}{*}{YOLO-W}
        & 15 & 100 &   0 &   0 &  0 & 0.0 \\
        & 20 & 100 &   0 &   0 &  0 & 0.0 \\
\bottomrule
\end{tabular}}
\end{table}

\begin{table}[t]
\centering
\caption{Generator comparison: B1/B2/B3 bucket distribution (\%) for
  llava-phi3 vs.\ moondream (T0=CC, $K{=}15$, stratified, 3 seeds).
  Bold B3 entries highlight configurations where moondream outperforms
  phi3 in keep rate. OWLv2 moondream runs not completed.}
\label{tab:generator-buckets}
\resizebox{\columnwidth}{!}{%
\begin{tabular}{ll|rrr|rrr}
\toprule
& & \multicolumn{3}{c|}{llava-phi3} & \multicolumn{3}{c}{moondream} \\
Backbone & Dataset & B1\% & B2\% & B3\% & B1\% & B2\% & B3\% \\
\midrule
\multirow{3}{*}{G-DINO}
  & COCO        & 67 & 22 & 11 & 73 & 24 &  3 \\
  & Cityscapes  &  0 & 98 &  2 & 64 & 36 &  0 \\
  & Chula Vista &  0 & 85 & 15 &  0 &100 &  0 \\
\midrule
\multirow{3}{*}{OmDet}
  & COCO        & 45 & 52 &  3 & 85 & 12 &  3 \\
  & Cityscapes  &  4 & 96 &  0 & 96 &  4 &  0 \\
  & Chula Vista &  0 & 98 &  2 &  0 & 97 &  3 \\
\midrule
\multirow{3}{*}{YOLO-W}
  & COCO        & 84 &  8 &  8 &  0 & 57 & \textbf{43} \\
  & Cityscapes  & 64 & 36 &  0 &  0 & 57 & \textbf{43} \\
  & Chula Vista &100 &  0 &  0 & 64 & 36 &  0 \\
\bottomrule
\end{tabular}}
\end{table}

\noindent{\bf Discussion:}
The results across all three experiments support a coherent narrative
about when and why caption refinement improves minority-class detection.

\noindent\textbf{Prompt sensitivity is the bottleneck, not detector capacity.}
Table~\ref{tab:caption-type} shows that the same frozen detector can
range from 0.00 to 33.18 minority AP@0.5 depending solely on prompt
choice (Grounding DINO/COCO).
C-GAP exploits this sensitivity systematically: by using detector
AP@0.5 as a feedback signal rather than requiring labels, it finds
better prompts than any static choice without touching model weights.

\noindent\textbf{Gains are largest where the CC baseline is weakest.}
The configurations with the highest absolute C-GAP gains are exactly
the cases where static captions completely fail to surface the minority
class.
This is by design: the method searches for prompts that shift detector
attention, and does so most visibly when that attention is currently
misdirected.
Figure~\ref{fig:qualitative-ex} illustrates this concretely: the CC
baseline misses Bike detections that C-GAP recovers using identically
specified frozen weights.

\noindent\textbf{The two-phase design is necessary.}
The $\mathcal{T}_0$ ablation (Table~\ref{tab:t0-ablation}) confirms
that LLM refinement alone is insufficient: starting from SD or CQ at
equal trial budget never recovers the gap to C-GAP from CC in the
primary experiments.
The composite caption is the necessary initialization, providing a
semantically complete foundation that triage can improve upon rather
than spending budget regenerating from impoverished starts.

\noindent\textbf{Generator--backbone pairing matters.}
Table~\ref{tab:generator-buckets} reveals that the moondream/YOLO-World
pairing achieves B3\,=\,43\% on COCO and Cityscapes, a result
invisible in the aggregate mean comparison (Table~\ref{tab:llm-comparison})
but critical for practitioners deploying on YOLO-class detectors.
Generator selection should be treated as a backbone-specific
hyperparameter, not a global choice.

\noindent\textbf{Triage behavior provides actionable diagnostics.}
B2-dominant configurations (OWLv2: B2\,=\,100\%) signal that the
initialization is near-optimal for the chosen margin $\delta$---either
the threshold should be widened or Phase~I is already close to the
detection optimum.
B1-dominant configurations (YOLO-World: B1\,=\,100\% on Chula Vista)
flag backbone--prompt incompatibility that warrants revision before
deployment.
These diagnostics are available without ground-truth labels, making
C-GAP useful as an evaluation tool even where AP@0.5 gains are limited.
\section{Conclusion}
\label{sec:conclusion}

We introduced C-GAP, a two-phase annotation-free framework that improves minority-class detection in frozen open-vocabulary detectors through iterative caption refinement.
Phase~I constructs per-image composite captions combining scene-level context with class-quantity grounding.
Phase~II uses detector-measured minority-class AP@0.5 as a label-free feedback signal to refine those captions via a three-bucket triage, requiring no parameter updates, no architectural changes, and no annotations at any stage.

Across four open-vocabulary backbones and three datasets spanning standard benchmarks and a real-world traffic-surveillance deployment, C-GAP improves minority-class AP@0.5 in 10 of 12 configurations, with gains up to $+14.40$\,pp on COCO and $+10.29$\,pp on the most severely imbalanced Chula Vista dataset.
Most strikingly, C-GAP recovers nonzero minority-class detections from zero-recall baselines purely by refining text prompts, with identical frozen weights, gains that are largest precisely where they matter most.

Beyond raw performance, the three-bucket triage exposes backbone--prompt
compatibility without ground-truth labels: B2-dominant configurations
signal a near-optimal initialization, while B1-dominant configurations
flag prompt-strategy mismatches that warrant revision before deployment.
This diagnostic value holds even when AP@0.5 headroom is limited, making
C-GAP useful as an evaluation framework independent of its refinement
gains.

Looking ahead, adaptive triage margins calibrated to per-backbone score variance would unlock gains on B2-saturated configurations, and extending the feedback signal to a multi-class minority objective would broaden applicability to general long-tail detection.
Most consequentially, deploying C-GAP as an online feedback loop, where prompts are continuously refined as a detector encounters new minority-class instances in the field, would bring annotation-free minority-class adaptation directly into production safety-critical systems.

\section{Endorsements}

Salimeh Sekeh has been partially supported by NSF CAREER CCF-2451457, and Arash Jahangiri and Francis Fernandez were funded, partially or entirely, by a grant from the Center for Pedestrian and Bicyclist Safety (CPBS), supported by the U.S. Department of Transportation (USDOT) through the University Transportation Centers program. The authors would like to thank CPBS and the USDOT for their support of university-based research in transportation. The findings are those of the authors only and do not represent any position of these funding bodies.

\FloatBarrier
{
    \small
    \bibliographystyle{ieeenat_fullname}
    \bibliography{main}
}
\clearpage

\twocolumn[{%
\begin{center}
  {\Large\bfseries Supplementary Material}\\[6pt]
\end{center}
}]

% \wacvssect{Supplementary Material}

This supplementary provides the extended results, dataset statistics, and implementation details that could not fit within the main paper's page limit.
Section~\ref{sec:extended-rw} expands the related work.
Section~\ref{sec:dataset-stats} reports per-class instance counts.
Section~\ref{sec:impl} gives full implementation details.
Section~\ref{sec:minority-only} reports the minority-class-only working-set probe.
Section~\ref{sec:llava} presents extended results for the LLaVA Phase~II generator.
Section~\ref{sec:generator-full} reports the full per-backbone generator comparison including bucket statistics.
Section~\ref{sec:caption-length} analyses caption length versus detection quality.

\section{Extended Related Work}
\label{sec:extended-rw}

\noindent\textbf{Class Imbalance in Object Detection.}
Class imbalance in detection has been comprehensively surveyed by~\cite{oksuz2020imbalance}, who taxonomize remedies into foreground--background imbalance (focal loss~\cite{lin2017focal}, repeat factor sampling~\cite{gupta2019lvis}) and foreground--foreground imbalance (equalization loss~\cite{tan2020equalization}, equalization loss v2~\cite{tan2021equalizationv2}, seesaw loss~\cite{wang2021seesaw}, class-balanced re-weighting~\cite{cui2019classbalanced}, logit adjustment~\cite{menon2021logit}, and decoupled classifiers~\cite{hong2021decoupling}).
Data augmentation remedies include copy-paste~\cite{ghiasi2021copypaste}, CutPaste~\cite{dwibedi2017cutpaste}, MixUp~\cite{zhang2018mixup}, CutMix~\cite{yun2019cutmix}, and mosaic tiling~\cite{bochkovskiy2020yolov4,cheng2024yoloworld}.
All require labeled minority examples and training-distribution intervention.
C-GAP acts at frozen-detector inference via prompt content, requiring no labels, no augmentation, and no retraining.

\noindent\textbf{Open-Vocabulary Detection.}
OVD decouples detection from fixed categories by grounding predictions in natural language~\cite{zareian2021ovrcnn,radford2021clip,li2022glip, li2022glipv2,kamath2021mdetr}.
Subsequent work introduced CLIP-based knowledge distillation~\cite{gu2021vild}, region-text contrastive pre-training~\cite{zhong2022regionclip}, vocabulary expansion via image-level supervision~\cite{zhou2022detic}, uncurated self-training~\cite{feng2022promptdet}, caption-bridged vocabulary transfer~\cite{bangalath2022bridging}, and unified detection--grounding architectures~\cite{shen2024ape,zang2022ovdetr}.
Group-level and compositional VL alignment is explored in~\cite{xu2022groupvit,yang2022unitab,yang2023improving}.
The four architectures we evaluate build on BERT~\cite{devlin2019bert}, CLIP~\cite{radford2021clip}, and self-supervised pre-training~\cite{caron2021dino,he2022mae}, with OWL-ViT~\cite{minderer2022owlvit} preceding the OWLv2~\cite{minderer2023owlv2} self-training scaling line, treat free-form text as the detection query so that prompt content directly controls what is detected without modifying any model weights.

\noindent\textbf{Prompt Engineering and Adaptation.}
CLIP~\cite{radford2021clip} established that template choice measurably affects zero-shot accuracy, motivating a rich body of soft prompt learning: CoOp~\cite{zhou2022coop}, CoCoOp~\cite{zhou2022cocoop}, KgCoOp~\cite{yao2023kgcoop}, MaPLe~\cite{khattak2023maple}, unsupervised adaptation~\cite{huang2022unsupervised}, and unified prompt learning~\cite{zang2022unified}.
Detection-specific learned prompts appear in DetPro~\cite{du2022detpro} and PromptDet~\cite{feng2022promptdet}.
Instruction-tuned VLMs, such as LLaVA~\cite{liu2023llava}, GPT-4V~\cite{openai2023gpt4v}, MiniGPT-4~\cite{zhu2023minigpt4}, Qwen-VL~\cite{bai2023qwenvl}, moondream~\cite{moondream2024}, and PaLI~\cite{chen2022pali}, are part of a broader family of multimodal instruction-following models~\cite{alayrac2022flamingo}.
C-GAP differs from all learned-embedding methods by using only discrete, human-interpretable captions, studying structural type (SD, CQ, CC) systematically and extending pseudo-caption paradigms~\cite{cho2023pseudocaption,chen2015cococaptions}.

\noindent\textbf{Iterative Refinement and Detector Feedback.}
Self-training pipelines~\cite{zareian2021ovrcnn,minderer2023owlv2, zhou2022detic,feng2022promptdet,xu2022groupvit,yang2022unitab, yang2023improving,bangalath2022bridging,kamath2021mdetr,xu2021softteacher} expand training from detector predictions but require parameter updates.
Test-time adaptation (TTP)~\cite{shu2022testtime}, entropy minimization~\cite{wang2021tent}, and surveys~\cite{liang2023ttasurvey}, adjust at inference without labeled targets.
LLM-driven feedback has improved captioning~\cite{cho2023pseudocaption, chen2015cococaptions}, visual reasoning~\cite{openai2023gpt4v, liu2023llava,zhu2023minigpt4,bai2023qwenvl}, and detection grounding~\cite{li2022glip,li2022glipv2,zhong2022regionclip, zhou2022detic,shen2024ape,zang2022ovdetr}.
C-GAP differs from all of the above by using aggregate minority-class AP@0.5, not per-sample confidence~\cite{shu2022testtime}, entropy~\cite{wang2021tent}, or distribution statistics~\cite{liang2023ttasurvey}, as the sole triage signal to selectively regenerate per-image captions via LLM, applying a structured three-bucket filter (keep, tentative, regenerate) with no detector parameter updates at any stage.

\section{Dataset Statistics}
\label{sec:dataset-stats}

\begin{figure*}[ht]
    \centering
    \includegraphics[width=\linewidth]{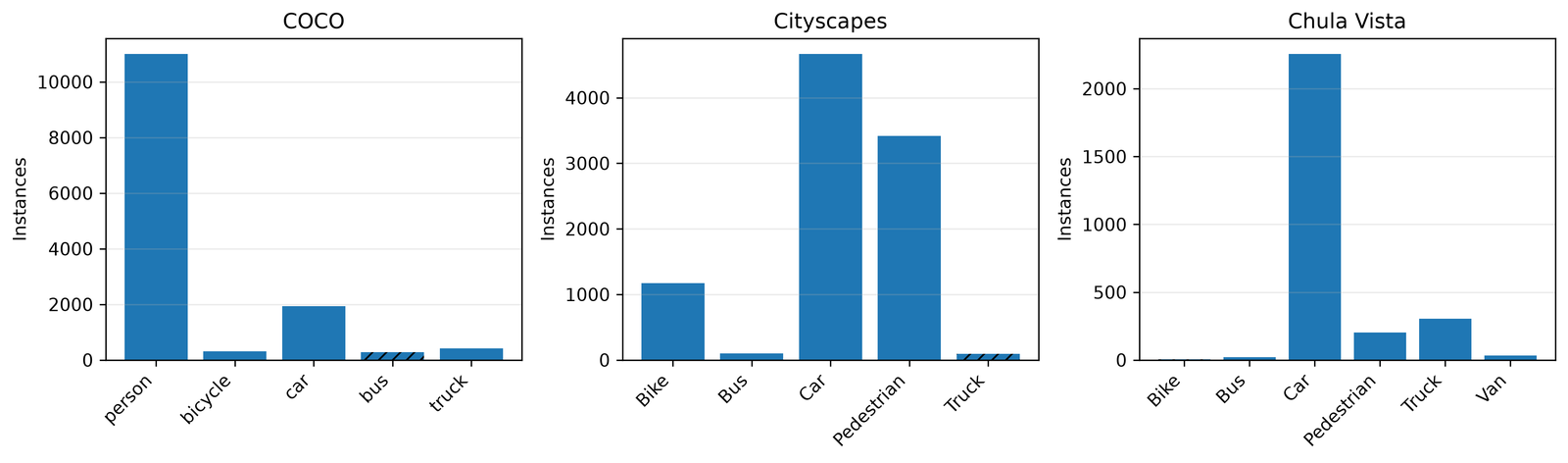}
    \caption{Class-instance distribution for the three evaluation datasets.}
    \label{fig:dataset_imbalance}
\end{figure*}

\begin{table}[h]
    \centering
    \caption{Per-class instance counts in each validation split.}
    \label{tab:dataset-stats}
    \resizebox{\columnwidth}{!}{%
        \begin{tabular}{lccc}
        \toprule
        Class & COCO (2948 img) & Cityscapes (500 img) & Chula Vista (288 img) \\
        \midrule
        Person / Pedestrian &  10,777 & 1,080 &   472 \\
        Bicycle / Bike      &   1,294 &   376 & 170 \\
        Car                 &   4,504 & 3,478 & 1,975 \\
        Bus                 &     283 &   152 &   226 \\
        Truck               &     615 & 204 & 303 \\
        Van                 &     --- &   --- &   316 \\
        \bottomrule
    \end{tabular}}
\end{table}

\begin{figure}[h]
    \centering
    \includegraphics[width=\linewidth]{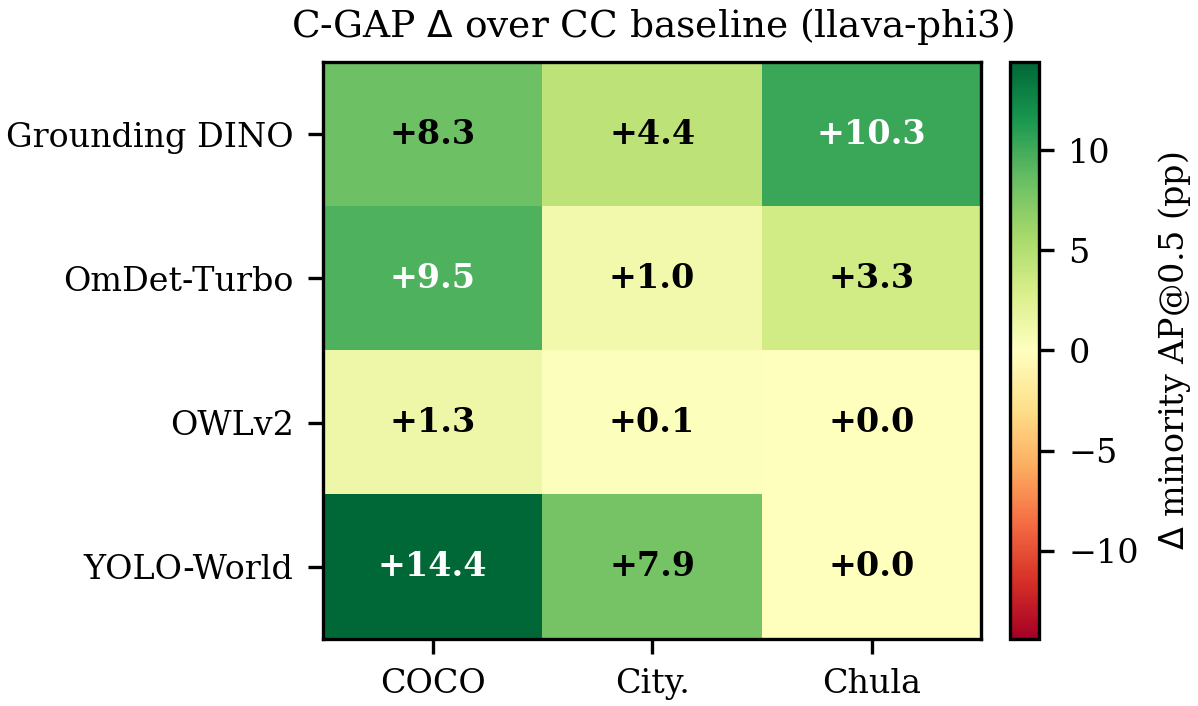}
    \caption{C-GAP minority AP@0.5 improvement ($\Delta$\,pp) over the CC baseline across all backbone--dataset pairs (llava-phi3, $K{=}15$). 
    Green cells indicate positive gains. 
    The single zero cell (OWLv2 and YOLO-World on Chula Vista) indicates C-GAP equals CC by construction. All improvements are driven purely by prompt refinement with frozen detector weights.}
    \label{fig:delta-heatmap}
\end{figure}

Table~\ref{tab:dataset-stats} reports per-class instance counts for the validation split used in each experiment.
The minority class in each dataset is italicized: bus in COCO (283 instances vs.\ 10,777 for person), truck in Cityscapes (204 instances vs.\ 3,478 for car), and Bike in Chula Vista (170 instances vs.\ 1,975 for Car). Figure~\ref{fig:dataset_imbalance} displays this imbalance across the datasets. Note the "Van" class doesn't exist for COCO or Cityscapes.

Figure~\ref{fig:delta-heatmap} provides a compact per-backbone-dataset view of all C-GAP minority AP@0.5 gains over the CC baseline.

\section{Implementation Details}
\label{sec:impl}

\subsection{Detector Inference Settings}

All detector weights are frozen throughout and no fine-tuning is performed at any stage. 
Each backbone uses its published default thresholds, which were not tuned on our datasets. 
All models run with batch size~8.

\noindent\textbf{Grounding DINO} (IDEA-Research/grounding-dino-base)
uses a box threshold of 0.30 and a text threshold of 0.25. 
Per-image captions are supplied as free-text queries alongside the fixed class list, and the model's BERT-based cross-modal fusion encodes both jointly.

\noindent\textbf{OmDet-Turbo} (omlab/omdet-turbo-swin-tiny-hf)
uses a score threshold of 0.30 and an NMS threshold of 0.50. 
Captions are passed as the free-text task prompt alongside the fixed class list, where OmDet-Turbo's efficient fusion head processes them together with the visual features.

\noindent\textbf{OWLv2} (google/owlv2-base-patch16-ensemble)
uses a score threshold of 0.10. 
Since the processor enforces a hard limit of 16 tokens per text query, captions are first tokenized and then chunked into segments of at most 14 tokens, leaving room for the BOS and EOS special tokens. 
Each chunk is appended as an additional text query alongside the fixed class names, and any detections attributed to a caption chunk are mapped back to the nearest class name via substring matching.

\noindent\textbf{YOLO-World} (yolov8l-worldv2.pt) uses a confidence threshold of 0.10. 
Captions are chunked into segments of at most 77 CLIP tokens to respect the vocabulary re-parameterization's input limit, and each segment is appended as an additional vocabulary entry alongside the fixed class list.

\subsection{C-GAP Protocol}

We use a maximum refinement budget of $K{=}15$, and bucket margin $\delta{=}0.05$ relative to the per-seed CC baseline AP@0.5.
On COCO, iterative refinement is performed on a 100-image stratified working set per run for tractability: at least 5\% of sampled images contain the minority class, and the remaining images are sampled from the rest of the filtered validation split.
Cityscapes and Chula Vista are evaluated on their full validation splits (500 and 288 images, respectively).
All comparisons within a run use the same working set.
Experiments were run on NVIDIA RTX~5080 and Ada Lovelace GPUs.

\noindent\textbf{LLM generation parameters.}
All Phase~II caption refinements use temperature 0.7 and a maximum output length of 256 tokens.

\noindent\textbf{Sequencing LLM and detector.}
After each trial's per-image LLM calls complete, the model is unloaded via \texttt{keep\_alive=0} and GPU-clear status is confirmed by polling \texttt{ollama~ps} (timeout 120\,s) before the detector subprocess is launched.

\noindent\textbf{Detector image restriction.}
When the per-image caption map covers only a subset of validation images (ex: the 100-image COCO working set), detector inference is restricted to exactly those images via stem-based filename matching against the annotation file, avoiding unnecessary computation on the remaining images.

\section{Minority-Class-Only Working Set}
\label{sec:minority-only}

The stratified working set used in the main experiments guarantees minority-class images are present but mirrors the natural class distribution of the full validation set, meaning roughly 95\% of the 100 sampled images contain no bus ground truth and therefore contribute zero signal to minority-class AP@0.5.

As a complementary probe, we run CGAP on the bus-containing images of the COCO validation split (all images with at least one bus ground-truth annotation, llava-phi3, $K{=}15$, $\mathcal{T}_0$=CC, three independent runs).
This isolates caption-refinement efficacy to scenes where the minority class is actually present.

Note that the CC baseline AP@0.5 differs between the two working-set regimes because the image populations differ: the minority-only set is evaluated entirely on bus-containing images, giving a higher baseline AP@0.5 than the stratified 100-image set.
Results across the two regimes are therefore not directly comparable in absolute terms.
The relevant comparison within each regime is CC vs. CGAP.

Table~\ref{tab:minority-only} reports CC and CGAP minority AP@0.5 under both regimes for all four backbones.
CGAP improves over CC on three of four backbones under the minority-only regime, confirming that the gains observed in the stratified experiments reflect genuine caption-refinement effects on images where the minority class is present.
OmDet-Turbo is the one exception: across all three runs the best trial is $k{=}0$ (the CC baseline itself), so CGAP equals CC.
This is consistent with the bucket statistics in Table~\ref{tab:minority-only-buckets}: OmDet-Turbo lands in B1 or B2 for every trial on the minority-only set, never exceeding $\tau_{\mathrm{high}}$.

\begin{table}[h]
    \centering
    \caption{Minority AP@0.5 (\%) on COCO/bus: stratified 100-image working set vs.\ minority-class-only (all bus-containing images), mean over three runs, llava-phi3, $K{=}15$.
      CGAP cannot score below CC by construction.
      CC baseline values differ between regimes because the image populations differ, and within-regime $\Delta$\,=\,CGAP$-$CC is the meaningful comparison.}
    \label{tab:minority-only}
    \resizebox{\columnwidth}{!}{%
    \begin{tabular}{l|ccc|ccc}
        \toprule
        & \multicolumn{3}{c|}{Stratified (100 img)} &
          \multicolumn{3}{c}{Minority-only (bus-containing)} \\
        Backbone & CC & CGAP & $\Delta$ & CC & CGAP & $\Delta$ \\
        \midrule
        Grounding DINO & 15.30 & 23.58 & $+$8.28  & 41.98 & 51.65 & $+$9.67 \\
        OmDet-Turbo    & 53.56 & 63.02 & $+$9.46  & 76.62 & 76.62 &    0.00 \\
        OWLv2          & 75.83 & 77.13 & $+$1.30  & 87.62 & 89.11 & $+$1.49 \\
        YOLO-World     & 17.69 & 32.09 & $+$14.40 & 20.47 & 28.40 & $+$7.93 \\
        \bottomrule
    \end{tabular}}
\end{table}

\begin{table}[h]
    \centering
    \caption{Bucket distribution and convergence on COCO/bus under the minority-only working set (llava-phi3, $K{=}15$, three runs), with stratified counterparts shown in parentheses for reference. Med.\,=\,median best-trial index; $\sigma$\,=\,std.\ dev.\ of per-run best-trial indices.}
    \label{tab:minority-only-buckets}
    \resizebox{\columnwidth}{!}{%
    \begin{tabular}{lcccccc}
        \toprule
        Backbone & B1\% & B2\% & B3\% & Med. & $\sigma$ &
          (Stratified B3\%) \\
        \midrule
        Grounding DINO &  0 & 40 & 60 & 1 & 1.2 & (11) \\
        OmDet-Turbo    & 27 & 73 &  0 & 0 & 0.0 &  (3) \\
        OWLv2          &  0 &100 &  0 &12 & 5.7 &  (0) \\
        YOLO-World     &  0 & 67 & 33 & 3 & 0.0 &  (8) \\
        \bottomrule
    \end{tabular}}
\end{table}

Table~\ref{tab:minority-only-buckets} reports the triage bucket distribution for all four backbones on the minority-only working set, alongside the corresponding stratified statistics for comparison.

Comparing minority-only and stratified bucket profiles reveals two notable shifts.
Grounding~DINO achieves B3\,=\,60\% at median best trial~1 on the minority-only set (vs.\ B1\,=\,67\%, B3\,=\,11\%, median~2 on stratified), indicating that concentrating on bus-containing images substantially increases the fraction of trials that exceed the keep threshold.
YOLO-World moves from B1-dominated on the stratified set (B1\,=\,84\%, B3\,=\,8\%) to B2/B3-dominated on the minority-only set (B2\,=\,67\%, B3\,=\,33\%), suggesting that when YOLO-World's vocabulary re-parameterization operates on a bus-dense image set, captions more often fall near or above the keep threshold.
OWLv2 remains B2\,=\,100\% in both regimes, consistent with the main paper's explanation: the CC initialization is near-optimal for OWLv2, so wider triage margins would be needed to trigger keep decisions.
OmDet-Turbo is the one backbone whose minority-only behavior is \emph{weaker} than stratified (B3 collapses from 3\% to 0\%, and all three runs converge at $k{=}0$), indicating that its high CC baseline on the bus-dense set leaves less refinement headroom rather than more.

\section{LLaVA as Phase~II Generator}
\label{sec:llava}

The main paper (Table~5) compares llava-phi3 and moondream as the two primary Phase~II generators. We additionally ran CGAP with LLaVA (7B) as an exploratory Phase~II refiner for all completed backbone--dataset configurations. Phase~I captions are held fixed at llava-phi3, with $K{=}15$, $\mathcal{T}_0{=}$CC, and the stratified working-set protocol. Since the LLaVA sweep was not completed for every backbone--dataset pair, these results are reported only as supplementary diagnostics and are not included in the main aggregate generator comparison.

\begin{figure}[H]
    \centering
    \includegraphics[width=\linewidth]{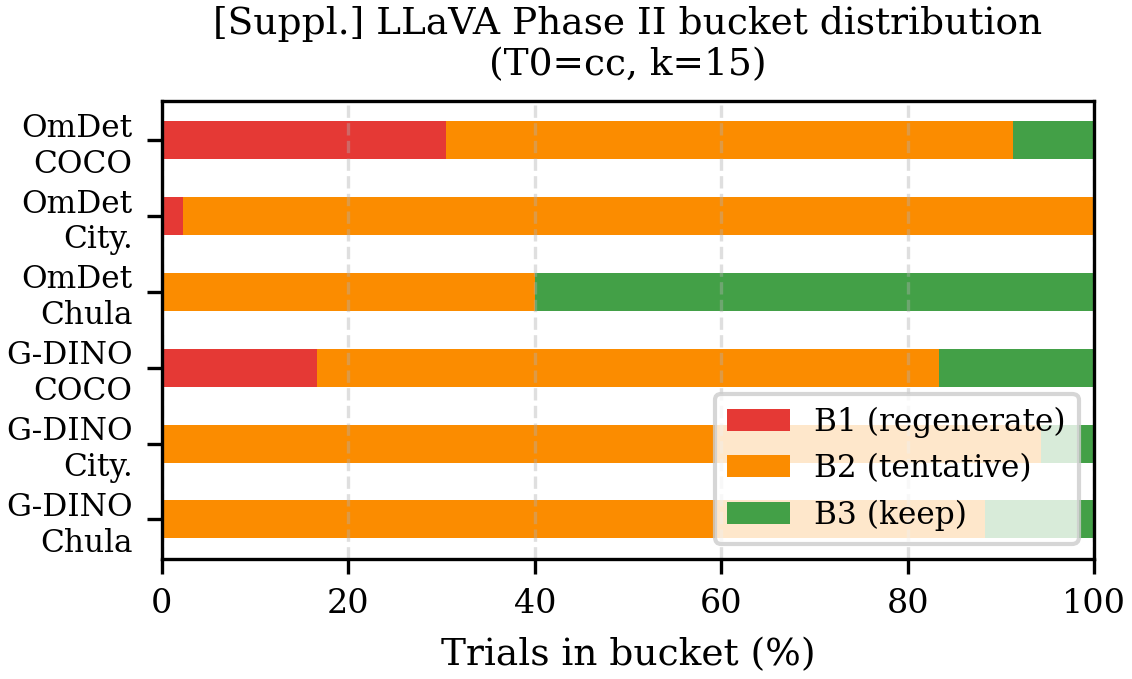}
    \caption{LLaVA Phase~II bucket distribution per backbone--dataset (T$_0${=}CC, $K{=}15$, stratified, three runs). Bars show the fraction of all trials falling into each bucket. B3 (keep) is non-zero for Grounding~DINO and OmDet-Turbo.}
    \label{fig:llava-bucket-dist}
\end{figure}

Table~\ref{tab:llava-buckets} reports the per-configuration bucket distributions for all completed LLaVA runs.
Figure~\ref{fig:llava-bucket-dist} visualizes these results. Several patterns emerge from Table~\ref{tab:llava-buckets}. First, LLaVA produces mostly B2 outcomes, indicating that its refined captions often remain close to the dynamic threshold rather than strongly exceeding or falling below the CC baseline. This behavior is most visible for Cityscapes and OWLv2, where B2 dominates nearly all completed runs. In these settings, LLaVA does not usually produce captions that are poor enough to trigger regeneration, but it also rarely produces captions that exceed the keep threshold. This suggests that the LLaVA refinements are generally conservative: they modify the prompt enough to remain competitive with the CC baseline, but not always enough to create a decisive minority-class AP@0.5 gain.

Second, the strongest LLaVA configuration is OmDet-Turbo on Chula Vista, where 60\% of trials enter B3 and the median best trial is $k=2$. This indicates that, for this backbone--dataset pair, LLaVA can quickly generate captions that improve over the composite-caption threshold. This behavior contrasts with OmDet-Turbo on Cityscapes, where 98\% of trials remain in B2 and none reach B3. The difference suggests that the usefulness of a Phase II generator depends not only on the generator itself, but also on how its caption style interacts with the detector architecture and dataset domain.

Third, LLaVA is poorly matched with YOLO-World in the completed runs. On COCO, all trials fall into B1, and on Cityscapes, 78\% of trials fall into B1 with no B3 trials. This implies that LLaVA refinements often reduce YOLO-World's minority-class AP@0.5 relative to the CC-centered threshold. Since YOLO-World uses a vocabulary re-parameterization mechanism rather than a separate long-context task prompt, this result is consistent with the broader generator analysis: longer or more descriptive VLM outputs are not always beneficial for detectors that are more sensitive to concise class-like vocabulary entries.

Overall, the LLaVA runs reinforce the main paper's conclusion that Phase II generator choice should be treated as a backbone-dependent design variable. LLaVA is not uniformly better or worse than llava-phi3 or moondream. Instead, it produces useful B3 trials for some configurations, especially OmDet-Turbo on Chula Vista, while showing B2 saturation or B1 failure modes elsewhere. Because the LLaVA sweep was not completed for every backbone--dataset pair, we report it as an additional diagnostic experiment rather than as part of the main aggregate generator comparison.

\begin{table}[H]
    \centering
    \caption{LLaVA Phase~II generator: bucket distribution per backbone--dataset (T$_0${=}CC, $K{=}15$, stratified, three runs). Med.\,=\,median best-trial index.}
    \label{tab:llava-buckets}
    \resizebox{\columnwidth}{!}{%
    \begin{tabular}{llcccr}
        \toprule
        Backbone & Dataset & B1\% & B2\% & B3\% & Med. \\
        \midrule
        \multirow{3}{*}{G-DINO}
          & COCO        & 17 & 67 & 17 & 7 \\
          & Cityscapes  &  0 & 94 &  6 & 7 \\
          & Chula Vista &  0 & 88 & 12 & 1 \\
        \midrule
        \multirow{3}{*}{OmDet}
          & COCO        & 30 & 61 &  9 & 3 \\
          & Cityscapes  &  2 & 98 &  0 & 3 \\
          & Chula Vista &  0 & 40 & 60 & 2 \\
        \midrule
        \multirow{2}{*}{OWLv2}
          & COCO        & 11 & 89 &  0 & 0 \\
          & Chula Vista &  0 &100 &  0 & 0 \\
        \midrule
        \multirow{2}{*}{YOLO-W}
          & COCO        &100 &  0 &  0 & 0 \\
          & Cityscapes  & 78 & 22 &  0 & 0 \\
        \bottomrule
    \end{tabular}}
\end{table}

\subsection{Comparison Across Generators}

Table~\ref{tab:generator-agg} places LLaVA alongside llava-phi3 and moondream, aggregated across all completed backbone--dataset combinations at $K{=}15$, $\mathcal{T}_0$=CC, stratified.

\begin{table}[H]
    \centering
    \caption{Aggregate bucket distribution by Phase~II generator ($K{=}15$, $\mathcal{T}_0${=}CC, stratified, all completed backbone--dataset combinations).}
    \label{tab:generator-agg}
    \begin{tabular}{lcccc}
        \toprule
        Generator & B1\% & B2\% & B3\% & $n$ trials \\
        \midrule
        LLaVA      & 29.7 & 66.6 & 3.7 & 323 \\
        llava-phi3 & 31.6 & 65.9 & 2.5 & 402 \\
        moondream  & 52.6 & 44.4 & 3.1 & 293 \\
        \bottomrule
    \end{tabular}
\end{table}

Several patterns emerge from Tables~\ref{tab:llava-buckets} and~\ref{tab:generator-agg}.

\noindent\textbf{LLaVA and llava-phi3 are comparable in aggregate.}
Both produce roughly 30\% B1 and 66--67\% B2 trials, with B3 rates below 4\%.
The similar profile reflects that both are instruction-tuned VLMs of broadly similar capability whose outputs sit near the triage thresholds.

\noindent\textbf{Moondream is more aggressive.}
Moondream triggers B1 in 53\% of trials, nearly double the rate of LLaVA or llava-phi3, because its shorter outputs consistently fall below $\tau_{\mathrm{low}}$ on Grounding~DINO and OmDet-Turbo.
The aggregate B3 rate (3.1\%) is similar, but this masks the strong YOLO-World inversion discussed in the main paper and detailed in Section~\ref{sec:generator-full}.

\noindent\textbf{OmDet-Turbo / Chula Vista is a strong configuration for LLaVA.}
LLaVA achieves B3\,=\,60\% at median best trial~2 for OmDet-Turbo on Chula Vista, matching the efficiency of the llava-phi3/$K{=}10$ result reported in the main paper (Table 6 OmDet $K{=}10$, B3\,=\,43\%, median~2).
This suggests OmDet-Turbo's efficient fusion head is particularly receptive to LLaVA-style scene descriptions on the Chula Vista intersection domain.

\section{Full Generator Comparison: Bucket Statistics}
\label{sec:generator-full}

Table~\ref{tab:generator-full} extends Table 5 of the main paper with the complete per-backbone, per-dataset bucket breakdown for llava-phi3 vs.\ moondream ($K{=}15$, $\mathcal{T}_0${=}CC, stratified, three runs).
The table covers the three backbones for which both generators were run on all three datasets: Grounding~DINO, OmDet-Turbo, and YOLO-World.

\begin{table}[H]
    \centering
    \caption{Generator comparison: B1/B2/B3 bucket distribution (\%) for llava-phi3 vs.\ moondream (T$_0${=}CC, $K{=}15$, stratified, three runs). \textbf{Bold} B3 entries highlight configurations where moondream achieves a higher keep rate than phi3.}
    \label{tab:generator-full}
    \resizebox{\columnwidth}{!}{%
    \begin{tabular}{ll|ccc|ccc}
        \toprule
        & & \multicolumn{3}{c|}{llava-phi3} & \multicolumn{3}{c}{moondream} \\
        Backbone & Dataset & B1\% & B2\% & B3\% & B1\% & B2\% & B3\% \\
        \midrule
        \multirow{3}{*}{G-DINO}
          & COCO        & 67 & 22 & 11 & 73 & 24 &  3 \\
          & Cityscapes  &  0 & 98 &  2 & 64 & 36 &  0 \\
          & Chula Vista &  0 & 85 & 15 &  0 &100 &  0 \\
        \midrule
        \multirow{3}{*}{OmDet}
          & COCO        & 45 & 52 &  3 & 85 & 12 &  3 \\
          & Cityscapes  &  4 & 96 &  0 & 96 &  4 &  0 \\
          & Chula Vista &  0 & 98 &  2 &  0 & 97 &  3 \\
        \midrule
        \multirow{3}{*}{YOLO-W}
          & COCO        & 84 &  8 &  8 &   0 & 57 & \textbf{43} \\
          & Cityscapes  & 64 & 36 &  0 &   0 & 57 & \textbf{43} \\
          & Chula Vista &100 &  0 &  0 &  64 & 36 &   0 \\
        \bottomrule
    \end{tabular}}
\end{table}

The bucket distributions reveal a clear generator--backbone interaction. For Grounding~DINO and OmDet-Turbo, llava-phi3 is generally more stable: most trials fall into B2, and B3 is observed in several configurations. In contrast, moondream often shifts these same backbones into B1, especially on COCO and Cityscapes, where 64--96\% of trials fall below $\tau_{\mathrm{low}}$. This suggests that moondream's shorter captions are often too sparse for detectors whose fusion mechanisms can benefit from richer scene and class-context descriptions.

YOLO-World shows the opposite pattern. With llava-phi3, YOLO-World is frequently B1-dominated, especially on Chula Vista where all trials fall into B1. With moondream, however, YOLO-World reaches B3 in 43\% of trials on both COCO and Cityscapes. This supports the interpretation that YOLO-World benefits from concise prompt entries because its vocabulary re-parameterization behaves differently from the task-prompt or cross-modal fusion mechanisms used by the other detectors. In this setting, shorter captions may reduce prompt dilution and make the minority class easier to surface.

The Chula Vista results further show that the moondream advantage is not universal even within YOLO-World. Although moondream reduces the B1 rate from 100\% to 64\%, it still produces no B3 trials on Chula Vista. This indicates that the generator effect depends jointly on the detector architecture and dataset domain. Concise captions help YOLO-World on COCO and Cityscapes, but do not fully overcome the more severe imbalance and visual difficulty of the Chula Vista Bike class.

Overall, Table~\ref{tab:generator-full} supports treating the Phase~II generator as a detector-specific design choice rather than a globally optimal component. llava-phi3 is the safer default for Grounding~DINO and OmDet-Turbo because it produces fewer below-threshold trials, while moondream can be preferable for YOLO-World because its concise outputs better match the detector's vocabulary-based prompting interface. The B1/B2/B3 statistics therefore provide a useful diagnostic: B1-heavy configurations indicate generator--detector mismatch, B2-heavy configurations indicate conservative refinements near the CC baseline, and B3-heavy configurations indicate settings where the generator frequently produces prompt sets that exceed the keep threshold.

\begin{figure}[h]
\centering
\includegraphics[width=0.92\linewidth]{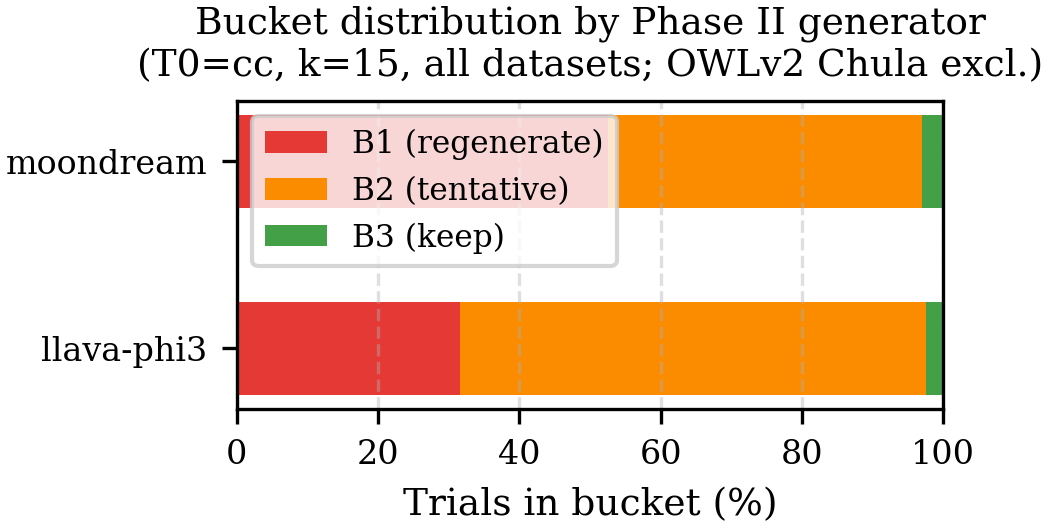}
\caption{Aggregate bucket distribution by Phase~II generator ($K{=}15$, $\mathcal{T}_0${=}CC, all available backbone--dataset combinations). Moondream's higher B1 rate reflects its shorter outputs falling below $\tau_{\mathrm{low}}$ on most backbones; its elevated B3 on YOLO-World (Table~\ref{tab:generator-full}) is diluted in this aggregate view.}
\label{fig:bucket-by-generator}
\end{figure}

Figure~\ref{fig:bucket-by-generator} summarizes the generator comparison aggregated across all backbone--dataset pairs at $K{=}15$, $\mathcal{T}_0$=CC, stratified.

\section{Caption Length vs.\ Detection Quality}
\label{sec:caption-length}

\begin{figure}[H]
\centering
\includegraphics[width=\linewidth]{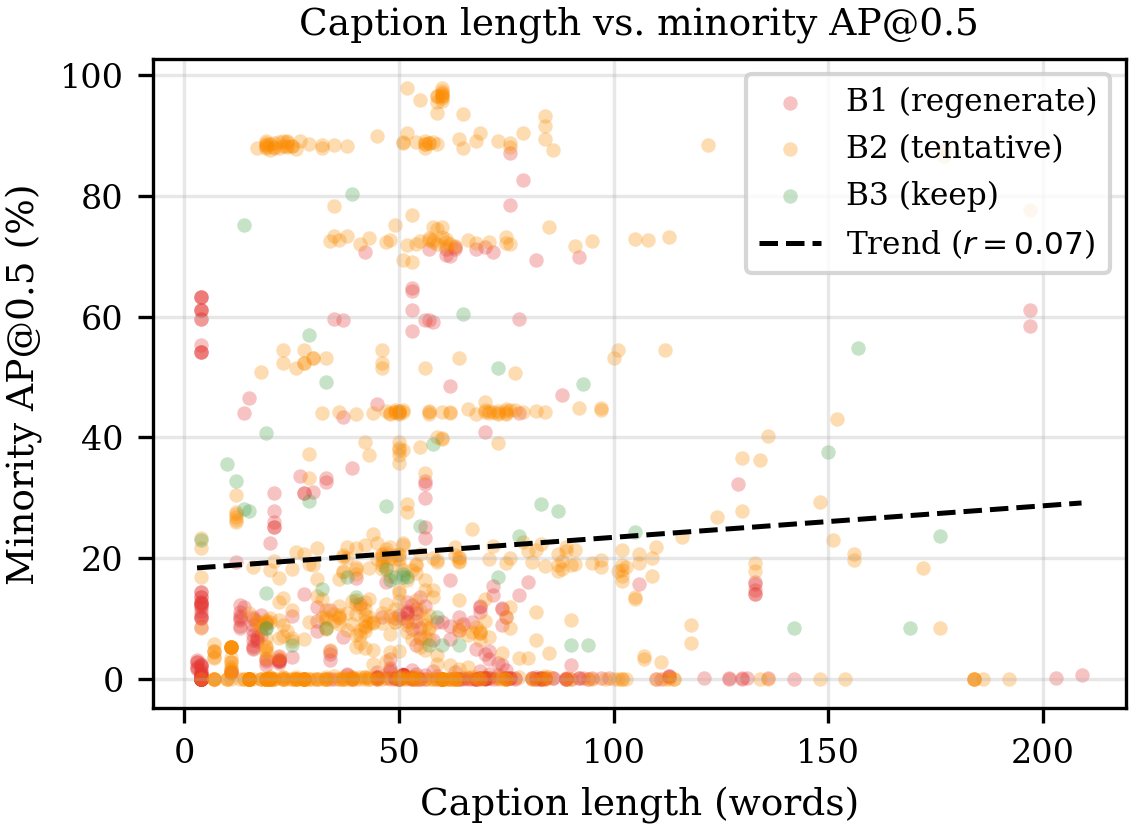}
\caption{Refined caption length vs.\ minority AP@0.5 across all CGAP trials ($n{=}1{,}104$, $r{=}0.07$). Points are colored by triage bucket. The near-zero correlation confirms that raw caption length is not a reliable proxy for detection quality.}
\label{fig:caption-length}
\end{figure}

Figure~\ref{fig:caption-length} plots refined caption length (words) against minority-class AP@0.5 across all 1,104 CGAP trials, colored by bucket assignment.
The Pearson correlation is $r{=}0.07$, indicating negligible linear dependence between caption length and detection quality.

Table~\ref{tab:caption-length-quartile} compares mean minority AP@0.5 for the shortest and longest caption-length quartiles.
Longest-quartile captions show a modest AP@0.5 advantage (20.6 vs.\ 16.2 for shortest), but the bucket distributions differ by only 1.1\,pp in B3 rate, confirming that caption length alone does not determine triage outcome.

\begin{table*}[ht]
\centering
\caption{Mean minority AP@0.5 and bucket distribution for the shortest and longest caption-length quartiles ($n{=}276$ each, all trials).}
\label{tab:caption-length-quartile}
\begin{tabular}{lccccc}
\toprule
Quartile & $n$ & Avg AP@0.5 (\%) & B1\% & B2\% & B3\% \\
\midrule
Shortest & 276 & 16.2 & 42.0 & 54.0 & 4.0 \\
Longest  & 276 & 20.6 & 37.3 & 57.6 & 5.1 \\
\bottomrule
\end{tabular}
\end{table*}

\section{Evaluation Protocol}
All detection outputs are stored in COCO-format JSON.
The primary metric is AP@0.5 for the designated minority class (bus/COCO, truck/Cityscapes, Bike/Chula Vista).
Secondary metrics are per-class AP@0.5 for all classes and overall mAP@0.5 across the full label set, computed with the standard COCO evaluation protocol (\texttt{pycocotools}).

For CGAP runs, the reported result for a given run is the highest minority-class AP@0.5 achieved across \emph{all} $k$ trials, including the Phase~I CC baseline trial ($k{=}0$).
This means CGAP cannot score below the CC baseline by construction.
All averaged results report the mean over three independent runs.

\subsection{Prompt Templates}
\label{supp:prompt_templates}

This section reports the exact prompt templates used to generate Phase I captions and Phase II C-GAP refinements. All prompts are discrete natural-language strings. No soft prompts, learned embeddings, or detector-parameter updates are used.

\paragraph{Scene Description Caption.}
For each image, the caption generator receives the image and the fixed class list. The prompt is:

\begin{quote}
Given the image and the following object classes: {class list}, describe the overall scene in detail, including the relationships between objects and any relevant context. Write a short caption in 1--2 sentences. Keep it concise and avoid long descriptions.
\end{quote}

\paragraph{Class Quantity Caption.}
The class-quantity caption is generated independently using the same image and class list:

\begin{quote}
Given the image and the following object classes: {class list}, estimate the number of visible instances for each present class. Do not guess objects outside the class list.
\end{quote}

\paragraph{Composite Caption.}
The initial C-GAP prompt is formed by concatenating the Scene Description and Class Quantity captions:
\begin{equation}
    t_{i,0} = \mathrm{concat}(t^{SD}_{i}, t^{CQ}_{i})
\end{equation}

\paragraph{Phase II Caption Refinement.}
At each refinement trial, the caption generator receives the image's original composite caption, the previous trial caption for the same image, the minority class, the class list, the previous aggregate AP@0.5, and the previous bucket assignment. The system instruction is:

\begin{quote}
You are refining a per-image caption used as a prompt for an open-vocabulary object detector. The detector receives exactly one caption per image and uses it for object detection on that image. Your job is to rewrite the caption to improve detection of the target minority class, while staying accurate to this image's actual content. Return only the rewritten caption. Do not explain your reasoning.
\end{quote}

The user prompt has the following structure:

\begin{quote}
CGAP per-image caption refinement task.

Allowed object classes: {class list}

Target minority class: {minority class}

This image's base description (scene description + object counts): {base caption}

Caption to refine: {previous caption}

{bucket-conditioned instruction}

Rewrite the caption to refine into ONE improved detector caption for THIS image.

Rules:
Return only one caption. Do not echo, restate, or refer to the rules or instructions. Do not use bullet points. Do not use markdown. Do not invent classes outside the class list. Base the caption only on the image's base description. If the base description does not mention the minority class, the refined caption must not introduce that class. If the base description already indicates the minority class is present, the generator may rephrase or emphasize that detail, but must not exaggerate its size, count, or prominence. Keep the caption concise but descriptive.
\end{quote}

\paragraph{Bucket-Conditioned Instructions.}
For B1, where the previous trial falls below the low threshold, the generator is instructed to improve minority-class recall and be more sensitive to small, distant, partially occluded, cropped, or side-view minority-class instances.

For B2, where the previous trial falls within the tentative range, the generator is instructed to retain the useful parts of the previous caption while improving wording for the minority class and balancing recall with precision.

For B3, where the previous trial exceeds the keep threshold, the generator is instructed to preserve the successful caption structure and produce only a small useful variation.

\end{document}